%% file: main.tex
\documentclass[10pt,leqno]{article}
\usepackage{tikz}
\usetikzlibrary{shapes, arrows, positioning}
\usepackage{graphicx}
\usepackage{indentfirst,csquotes}

\usepackage{amssymb,amsthm,amsmath}
\usepackage{xcolor,paralist,hyperref,titlesec,fancyhdr,etoolbox}

\newcounter{genplanxexample}[section]

\hypersetup{ colorlinks=true, linkcolor=black, filecolor=black, urlcolor=black }

\usepackage{verbatim}
\usepackage{xspace}
\usepackage{xcolor}
\usepackage{tcolorbox}
\usepackage{dirtree}
\usepackage{booktabs}
\usepackage{array}
\usepackage{geometry}
\usepackage{colortbl}
\usepackage{natbib}

\newcommand{\genplanx}{{\sc GenPlanX}\xspace}
\newcommand{\daniel}[1]{\textcolor{blue}{\textbf{Daniel says: }#1}}

\title{GenPlanX. Generation of Plans and Execution} 
\author{Daniel Borrajo \and Giuseppe Canonaco \and Tomás de la Rosa \and Alfredo Garrachón \and Sriram Gopalakrishnan \and Simerjot Kaur \and Marianela Morales \and Sunandita Patra \and Alberto Pozanco \and Keshav Ramani \and Charese Smiley \and Pietro Totis \and Manuela Veloso}
\date{JPMorganChase\\ \today}
\usepackage{listings}

\begin{document}
\include{PDDL}
\maketitle

\begin{abstract} Classical AI Planning techniques generate sequences of actions for complex tasks. However, they lack the ability to understand planning tasks when provided using natural language. The advent of Large Language Models (LLMs) has introduced novel capabilities in human-computer interaction. In the context of planning tasks, LLMs have shown to be particularly good in interpreting human intents among other uses. This paper introduces \genplanx that integrates LLMs for natural language-based description of planning tasks, with a classical AI planning engine, alongside an execution and monitoring framework. We demonstrate the efficacy of \genplanx in assisting users with office-related tasks, highlighting its potential to streamline workflows and enhance productivity through seamless human-AI collaboration. 
\end{abstract}

\section{Introduction}

The rapid advancement of AI has led to the development of techniques capable of understanding and executing complex tasks. Among these, Large Language Models (LLMs) have emerged as a powerful tool for interpreting natural language, enabling machines to comprehend and respond to human requests with remarkable accuracy~\cite{brown2020language}. However, the challenge remains in translating these requests into valid (and ideally optimal) plans that can be executed in real-world environments. 

In particular, we are interested on planning problems that involve the integration of standard office-related tasks, such as handling emails/calendars, managing presentations or databases, connecting to company APIs, or even running machine learning tasks. One of the pioneering efforts in this domain is the development of softbots, as introduced by Etizioni et al.~\cite{weld1992softbots}. These Softbots are software agents that perform tasks by interacting with software environments.

Currently, a preferred approach to solve these tasks consists of generating office co-pilots, by calling LLMs to generate plans in different formats~\cite{narayanaswamy2024usingcopilotMicrosoft}. However powerful such tools are to solve these tasks, a careful design of the LLM pipeline has to be defined given the lack of guarantees on correctness or optimality of these solutions.
Instead, classical AI planning offers robust methodologies for generating sequences of actions to achieve goals from given initial states~\cite{ghallab2004automated} for a specified domain model. AI planning has been shown to solve real-world tasks, such as logistics~\cite{or-timi}, satellite/rovers control~\cite{mapgen}, elevators control~\cite{koehler2000elevator}, or tourist plans~\cite{eswa-ondroad}. But, they lack the ability of ingesting natural language descriptions of tasks as required by current users. 

In this paper, we present the \genplanx system, GENeration of PLANs and eXecution, which is designed to receive requests in natural language about office-related tasks, generate plans to achieve the users intents, execute the actions in the generated plans, continuously monitor for successful execution, and replan in case of failed execution. Given a specific application in planning, the domain model is fixed and defined by humans. So, the users specify a planning task by providing the problem description. The problem is composed of a set of objects, an initial state and a set of goals (also called intents).  Furthermore, \genplanx is designed to allow seamless integration of new tools, making it adaptable to new applications. This is done by adding new actions to the PDDL domain description as well as their python code counterpart as explained later.
By integrating LLMs with classical planning techniques, we can create techniques that not only understand natural language requests, but also generate and execute plans that utilize various tools and databases to address these requests effectively.

\section{Related Work}

The integration of planning algorithms with Large Language Models (LLMs) has garnered significant attention in recent years, as researchers seek to enhance the capabilities of AI systems in understanding and executing complex tasks. This section reviews key contributions in this area, highlighting the advancements and challenges in combining these two technologies. The first subsection covers works that perform planning by fully using LLMs. The second subsection covers approaches where LLMs are used to help the planning process. The third subsection covers several frameworks for integrating LLMs and planning. And the fourth subsection describes planning approaches in the software domain.

\subsection{Planning using LLMs} 

Recent advancements in prompting strategies have enabled LLMs to internally orchestrate multi-step reasoning without relying on automated planning algorithms. One of the earliest approaches, Chain-of-Thought (CoT) prompting ~\cite{wei2023chainofthoughtpromptingelicitsreasoning} instructs the model to generate intermediate reasoning steps before arriving at a final output, effectively decomposing a complex task into simpler sub-problems. This method has proven instrumental in exposing the model's reasoning process and improving performance on tasks that require sequential decision-making. Building upon CoT, the Tree-of-Thought (ToT) methodology~\cite{yao2023treethoughtsdeliberateproblem} explores multiple reasoning paths concurrently, allowing the model to evaluate alternative approaches before selecting the most promising sequence of actions. These techniques enable the model to simulate a planning-like process by iteratively refining its output based on internal deliberation, despite the absence of explicit symbolic planning structures.

In addition to these reasoning paradigms, more complex techniques such as ReAct and ADaPT have been introduced to further enhance the internal planning capabilities of language models. ReAct ~\cite{yao2023reactsynergizingreasoningacting} interleaves reasoning steps with concrete actions in an interactive loop, allowing the model to incorporate real-time feedback from the environment into its planning process. ADaPT~\cite{prasad2024adaptasneededdecompositionplanning}, on the other hand, focuses on adaptive decomposition, where the model dynamically adjusts its strategy by breaking down tasks into smaller, manageable parts based on the context and observed performance. These methods highlight a new trend of LLM-based planning where internal guiding mechanisms and iterative self-correction are used in place of external planners. These techniques have been shown to handle open-ended planning tasks when there is no explicit domain model. However, they do not provide any guarantees in terms of soundness (valid solutions) or optimality, as can be expected from classical AI planning approaches. 

Unlike these approaches, \genplanx does not use LLMs for generating the plan. It relies on a classical planner to do so.

\subsection{LLM and Automated Planning}

Initial attempts to use LLMs for planning tasks through direct prompting have highlighted significant limitations~\cite{valmeekam2022large,rao_cannot_plan}. These studies demonstrated that LLMs alone struggle to generate valid plans when evaluated against standard planning benchmarks. These findings underscore the need for hybrid approaches that go beyond only using the capabilities of LLMs, such as combining LLMs with automated planners.

Recent research has shown that combining LLMs with external planners or validators can significantly enhance planning outcomes~\cite{LLM_modul_plan_Rao,arxiv-trip-pal}. This hybrid approach aligns with our work in developing \genplanx, where we leverage the strengths of both LLMs and classical planning techniques. 
In such approaches, an LLM parses the input request and generates a structured representation of the intent, while dedicated planners (such as those that support PDDL, the standard Planning Domain Description Language~\cite{PDDL}) are in charge of solving the planning problems formulated by LLMs. Such approaches were used in~\cite{guan2023leveraging,liu2023llm+}. LLMs have been shown to generate PDDL problems and domains reliably~\cite{oswald2024large,pallagani2024prospects}. The synergy between LLMs and planners enhances the inference-time behavior of LLMs and is at the core of our \genplanx architecture. In the case of \genplanx, instead of asking the LLM to generate a PDDL problem definition, we ask the LLM to generate a type of output that is easier for them to generate, as it is the case of json (or python dictionaries). Also, PDDL does not allow for the representation of objects that go beyond symbols, as paths to files or email addresses. Since they are needed for executing the actions of the plan, the output dictionary contains such information, as explained later.

Other work has focused on integrating structured domain knowledge with neural architectures.~\cite{gupta2021transformer} and~\cite{zhang2022structured} demonstrated that embedding explicit domain rules within transformer models improves the mapping accuracy from natural language to formal planning constructs. By aligning the semantic space of natural language with that of symbolic representations, these methods reduce ambiguities and enhance the robustness of the resulting plans. Analogously, \genplanx aligns the language and intents of the user to a formal planning representation (as opposed to the other way around), thus enhancing the robustness of the plan output, and providing plan guarantees.

\subsection{Frameworks for Integrating LLMs and Classical Planning}

Developing frameworks that facilitate the seamless integration of LLMs and planning algorithms is a critical area of research.~\cite{huang2022language} presented a framework that leverages LLMs to generate high-level plans, which are then refined by classical planning algorithms to ensure feasibility and efficiency. In a similar manner,~\cite{liu2023llm+} presented the LLM+P framework, in which an LLM first translates a natural-language task description into a PDDL problem specification, and then a classical planner is invoked to compute an optimal plan. Finally, the plan is translated back to a readable form. These approaches underscore the complementary strengths of LLMs in understanding language and planning algorithms in optimizing task execution.

Additionally,~\cite{mottaghi2020nocturne} introduced a framework for integrating LLMs with planning in interactive environments, enabling systems to execute complex tasks by combining language understanding with strategic planning. This work demonstrates the potential of such frameworks in applications ranging from robotics to virtual assistants.

Another hybrid strategy is to use LLMs for task decomposition and guidance while leaving low-level planning/execution to other methods.~\cite{singh2024twostepmultiagenttaskplanning} proposed TwoStep for multi-agent planning, where an LLM decomposes a goal into independent sub-goals for each agent, effectively assigning tasks in parallel. The LLM’s commonsense reasoning helped split the problem in ways a PDDL solver would not know to do, while the symbolic planners handle the sub-problems optimally.

\subsection{Softbots and Planning}

Most real-world applications of AI planning technology are related to control of physical systems, such as robots~\cite{mapgen} or satellites~\cite{muscettola98remote,ai-magazine}. Significantly less work has focused on software actions.
Early work~\cite{weld1992softbots} introduced the concept of softbots (software bots), which utilized planning to interact with software environments.
Among other works that have used AI planning for software tasks where actions are functions/commands to be executed in a computer we can mention web service composition~\cite{traverso2004automated,aicomm05}, business workflows generation~\cite{expertsystems07}, networks~\cite{patra2021using} or machine learning workflows\cite{mldm09}.

\section{Architecture}
\label{sec:architecture}

The architecture of \genplanx is designed to seamlessly integrate natural language processing with classical planning to automate complex tasks. \genplanx consists of several key components, each playing a crucial role in processing requests and executing plans. Figure~\ref{fig:architecture} illustrates the architecture of \genplanx.  There are two types of components in the architecture. Components in green are unique implementations for our approach, and contributions of this paper. Components in white are existing AI tools that were developed by others. Before we present each component, we briefly define the input that triggers an LLM+planning cycle, the user request. A request from a user contains several types of information:

\begin{itemize}
    \item Entities: objects that are relevant for the planning task. They will be translated into PDDL objects. Examples are: file names, titles of slides, appointments of a calendar, ...
    \item Initial state: starting configuration of the task. Examples are: a csv file contains a database; or some text was provided by the user to be translated.
    \item Goals (also called intents, or objectives): what the user would like the generated plan to do. Examples are: generating a PowerPoint with some specified contents; or creating a decision tree from the data initially stored in a csv file. 
\end{itemize}

Given that users do not need to know all these elements of AI planning, and that users instead describe the task in natural language, we need a component of the architecture that can map english descriptions of tasks into formal structured representations of planning tasks. This is why we incorporate an LLM-based component.
An overview of each component is as follows:

\begin{figure}[hbt]
    \centering
    \includegraphics[width=\textwidth]{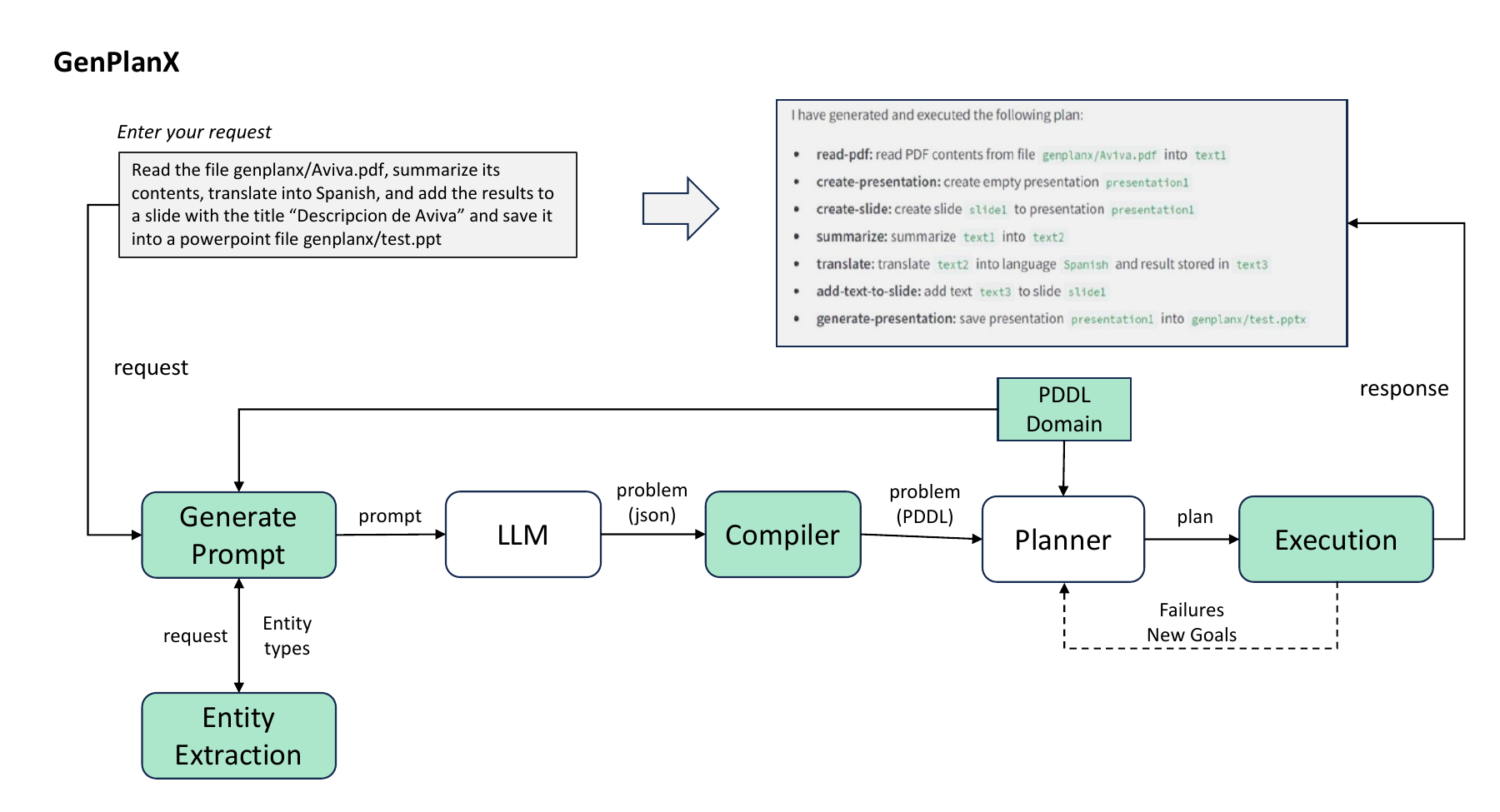}
    \caption{Architecture of \genplanx. Boxes in green are the unique implementations for our approach. Boxes in white are components integrated from existing AI tools.}
    \label{fig:architecture}
\end{figure}

\begin{itemize}
    \item \textbf{Entity Extraction}: This module analyzes the incoming request to identify relevant entities and their types. These entities and entity types are part of the organizational knowledge and are usually not known by general-purpose LLMs. This step is essential for understanding the context and specifics of the request, enabling \genplanx to generate accurate prompts for further processing.
    \item \textbf{Generate Prompt}: Once the entities and types are extracted, the `Generate Prompt' component constructs a prompt based on the user request, the entities detected, and the planning domain model. The objective of the Prompt Generation module is to provide a dynamic prompt that includes the context of the LLM task and some examples of how to assemble a structured response depending on the user intent. 
    \item \textbf{LLM}: This component handles the call to an LLM using the prompt generated previously. This task leverages the capabilities of LLMs to interpret the prompt and produce a structured problem description in json format. This includes the set of objects, and the initial state and  goals.
    \item \textbf{Compiler}: It translates the problem description from the json format into PDDL (Planning Domain Definition Language)~\cite{PDDL}, which is a standard language used in AI planning. This translation is crucial for enabling the Planner to generate executable plans.
    \item \textbf{Planner}: This component represents an AI Planning solver, which uses the PDDL problem description together with the domain description in PDDL to generate a sequence of actions, a plan, that addresses the request. The plan is guaranteed to achieve the goals from the initial state by using the available actions defined in the domain model.
    \item \textbf{Execution}: This module carries out the actions specified in the plan. It monitors the execution process, checking for successful completion of each action. In the event of execution failures or when new goals appear by the execution of previous actions, \genplanx can replan to generate a new plan.
    \item \textbf{Response Generation}:  Finally, \genplanx generates a response based on the execution results, providing feedback to the user. This response includes details of the executed plan and any outcomes obtained.
\end{itemize}

In the following sections, we will go over each component in detail.

\section{Entity Extraction}
\label{sec:entities}

The entity extraction module builds upon the capabilities outlined in~\cite{2025-AMP}. The work presented in the paper described an email handling solution. We used it within \genplanx to enhance the prompts in the presence of domain-specific entities that LLMs were not trained on. This is particularly crucial in the finance industry, where entities extracted include unique identifiers (for teams, firms, clients), security IDs (CUSIP, SEDOL, ISIN), trade economics (volume, amount, currency, dates), portfolio IDs, and account numbers among other entity types. 

Utilizing an ensemble approach that combines deep learning, pattern-based techniques, and domain expertise, the module effectively extracts entities from text. For instance, in case the input request contains references to entities such as {\tt CIDTA12}, {\tt F34GP5}, {\tt US1234567892}, {\tt 16-07-24}, {\tt A12345}, and {\tt P6763} the module categorizes them correctly as {\tt client identifier}, {\tt firm identifier}, {\tt ISIN}, {\tt trade date}, {\tt account number}, and {\tt portfolio id}, respectively. It returns a structured output detailing the types and lists of entities found, such as: {\tt \{`client identifier': [`CIDTA12'], `firm identifier': [`F34GP5'], `ISIN': [`US1234567892'], `trade date': [`16-07-24'], `account number': [`A12345'], `portfolio id': [`P6763']\}}.

\section{Domain Model}
In our work, we have named the domain as \texttt{assistant}, and it is designed 
to handle a variety of tasks within a professional office environment such as data manipulation, presentation, and communication. 
A PDDL domain model defines the knowledge of a planning environment by specifying key elements such as types, predicates, and actions. Types categorize objects within the domain, while predicates describe properties and relationships among these objects. Actions are defined by their preconditions and effects, detailing the operations that can be performed and how they change the state of the world. 

The \texttt{assistant} domain includes a diverse set of types, such as \texttt{file}, \texttt{dataframe}, \texttt{graph}, \texttt{data-type}, \texttt{email}, \texttt{contents} or \texttt{object}.  The domain also defines predicates such as \texttt{(in ?c - contents ?c1 - contents)} and \texttt{(available ?o - object)}, that are used to track which contents is associated with each other and to determine the availability of objects within \genplanx, respectively. Additionally, predicates such as \texttt{(used ?c - contents)} help prevent the creation of duplicate contents by marking them as utilized, while \texttt{(data-type-contents ?dt - data-type ?dc - data-contents)} provide insights into the specific data types that contents represent.

Among the various actions defined in this domain, we present as an example the \texttt{read-data} action which is defined as in Listing~\ref{lst:add-to-slide-action}. This action reads the data available in a \texttt{data-file}, which could be either an \texttt{excel-file} or a \texttt{csv-file}, and makes it \texttt{available} as a dataframe \texttt{?d}. The only precondition is that the dataframe contents is stored in the file.We also define a cost of 1 for this action.

\begin{lstlisting}[
  float=!htb,
  caption={{\tt read-data} action.},
  label={lst:add-to-slide-action},
  language=PDDL]
(:action read-data
   :parameters (?a - ai-agent ?d - dataframe ?f - data-file)
   :precondition (and (in ?d ?f))
   :effect (and (available ?d)
                (increase (total-cost) 1)))
\end{lstlisting}

\section{Prompting and LLMs} 
\label{sec:prompting} 
In this section, we explore the structure of the input prompt, detailing all the components necessary for \genplanx to effectively fulfill user requests. This includes the required intents, as well as the structure of the output, supplemented with examples.

\subsection{Prompt Description} 

\genplanx must interpret a given request, 
and identify a set of intents within the request. The prompt used by \genplanx is a carefully crafted set of instructions designed to assist in generating responses to office-related queries which can utilize structured data from multiple systems of records (SORs), understand user intents, and provide the necessary initial and goals for the planner. The prompt encompasses several key components, each serving a distinct purpose to ensure the response process is both efficient and accurate. We leverage the robust few-shot learning capabilities of LLMs to adapt \genplanx to several user requests. Below is an overview of the prompt-components, along with the rationale for their inclusion:

\subsubsection{Generic Task Definition}
The prompt begins by defining the generic task: responding to queries related to office operations. This establishes the context and scope of the task, which helps the LLM understand its primary objective. By clearly stating the task, the prompt provides a focused framework for generating relevant responses. Although the current configuration put in the prompt is static, the Model Context Protocol (MCP)\footnote{https://modelcontextprotocol.io/} can be leveraged to dynamically populate this and other sections of the prompt. Such prompt construction would allow for the automatic integration of tools and actions based on upto-date information, enhancing the flexibility and adaptability of the system in addressing diverse user queries. In the example we have provided in~\ref{app_example_prompt}, this task definition is illustrated by \textit{\quote You are working in an office environment. You have to provide responses to queries from employees or clients related to sales.\endquote}

\subsubsection{Database Schemas} The prompt provides schemas for relevant SORs in case the request refers to the available datasets. These can be databases pertaining to several office processes such as sales, operations, communications, etc. Each schema lists the relevant columns, such as \texttt{trade-id}, \texttt{client-id}, \texttt{date}, etc. The prompt instructs the LLM to pay attention to the case sensitivity of these fields when creating database queries. These schemas outline the available data fields, guiding the LLM in selecting the appropriate data sources for each query. This ensures that responses are based on accurate and relevant data. In the example we have provided in~\ref{app_example_prompt}, this task definition is illustrated by \textit{\quote Information is present on several systems of records (SOR). The schemas (columns) of those systems and their values are the following:
Schema for SOR 1: [list of columns], SOR 2: [list of columns] \endquote}

\subsubsection{Domain Types, Actions and Predicates} The prompt includes a complete list of domain types, predicates and actions by including all these elements from the input domain model in PDDL.  Including these ensures that the LLM can accurately model the data and actions required to fulfill the query. Several types like \textit{pie-chart, bar-chart} are mentioned in~\ref{app_example_prompt}, in addition to predicates like \textit{in-graph, in-data, etc.}

\subsubsection{Set of Intents} A set of predefined intents is included to guide the identification of user requests; this is to facilitate few-shot learning by the LLM. By categorizing requests into specific intents, the prompt helps streamline the response generation process, ensuring that each query is addressed appropriately. From the example in Appendix ~\ref{app_example_prompt}, we see that the section starting with \textit{Intent: \textless intent\textgreater\ Output:\textless dictionary\textgreater} exemplifies this.

\subsubsection{Response Format and Dictionary Structure}

The response to a query is expected to be a Python dictionary that encapsulates the identified intents. The dictionary must include keys for \texttt{init\_state} and \texttt{goals}, both of which are mandatory. The rest of the keys represent the objects that should be considered when solving the planning task. The values of all keys are definition dictionaries that include specific elements such as \texttt{type}, \texttt{value}, and other context-specific keys (e.g., \texttt{to}, \texttt{body}, or \texttt{subject} for emails). These structural constraints in the response ensure consistency and standardization in the output, making them easier to interpret and process.

In Appendix~\ref{app_example_prompt}, we see this exemplified by the following instructions:
\begin{quote}
Then, return a Python dictionary that contains information on all intents. You should not define a function or provide Python code, but return the dictionary as your output. Do not use external tools. The keys of the dictionary are the task entities, the \texttt{init\_state} and the \texttt{goals}.
\end{quote}

\subsubsection{State Representation}

The prompt also includes some definitions related to both \texttt{init\_state} and \texttt{goals}. They should be strings formatted as a sequence of literals. Each literal is a tuple with elements separated by spaces, where the first element is a predicate from the predefined list of predicates in the domain. The prompt also specifies that literals from the goals should not be included in the initial state. Appendix~\ref{app_example_prompt} contains the following statement, which is one way of expressing this:
\begin{quote}
Everything that is true at the start should be in the \texttt{init\_state} ... Do not include literals from the goals in the initial state.
\end{quote}

\subsubsection{Merging Intents} If multiple intents are identified in a request, the prompt instructs to merge the dictionaries into a single comprehensive dictionary. This involves combining all entities found and merging the \texttt{init\_state} and \texttt{goals} from all intents. One way of expressing this as seen from~\ref{app_example_prompt} is:

\textit{\quote If you find more than one intent in the request, merge the dictionaries into a single dictionary... \endquote}

\subsubsection{Other Constraints} The prompt also imposes constraints, such as not defining functions or using external tools, and ensuring that the output is formatted correctly in a single response.

A complete example prompt is provided in~\ref{app_example_prompt}. This example serves as a practical reference for users, demonstrating how to implement the guidelines effectively.

\subsection{Intents}

The intents refer to the set of goals that \genplanx should achieve to fulfill the user request. Each intent is a tuple composed of the name of the intent, its description and the expected json output from the LLM. These intents enable \genplanx to efficiently handle file and data management tasks, such as reading and saving files in various formats, including PDFs and Word documents, and performing database operations like adding, deleting, or modifying records. \genplanx also supports data visualization and presentations creation, enabling users to generate charts and comprehensive PowerPoint presentations with ease. Additionally, it facilitates effective communication by allowing users to send emails and notifications, ensuring that important information is shared promptly.

Beyond basic data handling, \genplanx offers advanced information processing and organizational tools. Users can explain, translate, or summarize text, find information within files, and conduct deep research on specific topics. \genplanx also aids in scheduling by identifying free slots in calendars and provides web search capabilities for additional information. For more complex interactions, it can query a large language model for intelligent responses and match files based on specific criteria, showcasing its versatility and adaptability in addressing a wide range of tasks. The following is the specific list of all current intents:

\begin{itemize}
    \item \textbf{File manipulation:} Read file, Save file, Read PDF file, Read Word file, Save PDF file.
    \item \textbf{Database primitive operations:} Add a record, Delete records, Count, Modify records, Add value.
    \item \textbf{Office-related operations:} Send email, Notify by email, Generate Chart, Create PowerPoint, Create chart slide, Create text slide, Create table slide.
    \item \textbf{LLM-related operations:} Explain, Translate, Summarize, Deep research, Ask LLM.
    \item \textbf{Other:} API access, Data access, Find information in file, Find free slots, Search web, Match files.
\end{itemize}

\subsection{Output of the LLM}

The output of the LLM is a Python dictionary/json that should contain the required information to fulfill the user request. This dictionary is composed of the joint information from the different set of intents selected by the LLM. It contains information on objects to be considered in the planning episode, the initial state representing the state at the beginning of the planning task, and the goals representing the desired partial state. This dictionary must adhere to the following structure:

\begin{itemize}
    \item The keys of the dictionary are the elements of the task (objects), \texttt{init\_state}, and \texttt{goals}. All keys must be in lowercase.
    \item Each object is defined as a dictionary with at least the two keys {\tt type} and {\tt value}. {\tt type} represents the PDDL type of the object. {\tt value} is the value that the object will take at execution time. Initially, it can have any arbitrary value for most cases. When actions operate with an object, the value of the object during execution will be saved there. Additionally, it can have other specific keys such as \texttt{to}, \texttt{body}, or \texttt{subject} for emails.
    \item The types of the different objects must adhere to a pre-defined set of types, the ones provided as input in the prompt that appear in the domain file.
    \item \texttt{init\_state} and \texttt{goals} are mandatory keys and  are composed of a sequence of literals. 
\end{itemize}

As an example, given the request {\it What is the status of the trade TR123?}, the LLM could return the dictionary in Figure~\ref{fig:output-dict}. This dictionary represents the output of the LLM tasked with extracting the status of a particular trade when provided with the trade ID, which is present in a downstream file. The file is identified by the type of the entity returned by the Entity extraction module. Going into more detail, what the LLM is describing at the output is:

\begin{figure}[htb!]
\centering
\begin{verbatim}
{"data-file1": {"type": "data-file", "value": "./genplanx/file_1.csv"}
 "dataframe1": {"type": "dataframe", "value": []}
 "filtered-dataframe": {"type": "dataframe", "value": []}
 "chat-response": {"type": "response", "value": []}
 "query1": {"type": "query", "value": "df[(df["trade-id"] == "TR123")]"}
 "init_state": {"type": "state", 
                "value": "(in dataframe1 data-file1) (available query1)
                          (query-result dataframe1 query1 filtered-dataframe)"}
 "goals": {"type": "state", 
           "value": "(and (done-query query1)
                          (in filtered-dataframe chat-response)
                          (sent chat-response))"}}
\end{verbatim}
\caption{Example of a structured output dictionary for getting the status of a trade.}
\label{fig:output-dict}
\end{figure}

\begin{itemize}
    \item \textbf{Objects}: The dictionary includes the elements (as key-value pairs) that are required during execution, like the file where the information is stored, the dataframe that will be extracted from the file, the filtered-dataframe after filtering it regarding the user query and the chat response that would be generated and returned to the user
    \item \textbf{Initial State}: Represents the initial state of the task, where the dataframe is inside the file, the system has the query defined and there is a query result after applying the query to the dataframe inside the file.
    \item \textbf{Goals}: Represents the desired state where the query has been done, the filtered dataframe is present in the response to the user and the response is sent to the user with the information requested.
\end{itemize}

\section{Planning and Execution}

In this section, we will cover the planning and execution of the generated task by the LLM. Given the dictionary returned by the LLM, the compiler module translates its information into a PDDL problem definition. As an example, the previous dictionary would be converted into the problem represented in Listing~\ref{list:problem}.

\begin{lstlisting}[
  float=!htb,
  caption={Problem Example.},
  label={list:problem},
  language=PDDL]
  (define (problem test-llm)
    (:domain assistant)
    (:objects ai - agent
              data-file1 - data-file
              dataframe1 filtered-dataframe - dataframe
              chat-response - response
              query1 - query)
    (:init (in dataframe1 data-file1)
           (available query1)
           (query-result dataframe1 query1 filtered-dataframe))
    (:goal (and (done-query query1)
           (in filtered-dataframe chat-response)
           (sent chat-response))))
\end{lstlisting}

\subsection{Planning}

The domain and problem files serve as input to the architecture described in Figure~\ref{fig:architecture}.
The Planner then returns a Plan, i.e., a sequence of actions described in PDDL that achieve the goal stated in the problem. \genplanx uses Fast-Downward ~\cite{helmert2006fast} through the Unified Planning library~\cite{unified_planning_softwarex2025}, but any other planning engine can be used, since the domain and problem definitions are specified in the PDDL standard language. Thus, \genplanx is planner-independent. In Listing~\ref{lst:plan}, we show an example of a plan in PDDL, solution to the problem stated in~\ref{list:problem}. Every action is represented as a tuple composed of the name and the parameters of the action.

\begin{lstlisting}[
  float=!htb,
  caption={Plan Example.},
  label={lst:plan},
  language=PDDL]
  (read-data ai dataframe1 data-file1)
  (query-data ai query1 dataframe1 filtered-dataframe)
  (create-response ai chat-response)
  (add-to-response ai filtered-dataframe chat-response)
  (send-response ai chat-response)
\end{lstlisting}

\subsection{Execution and Monitoring}

This plan is sent to the Monitoring and Execution module, which  translates it into the actual code that will be executed into the real-world. This translation is accomplished through a mapping linking PDDL actions to the corresponding python executable functions. Each PDDL action has a corresponding python function with the same name that implements it. The python implementation corresponding to the action represented in Listing~\ref{lst:add-to-slide-action} is shown in Listing~\ref{lst:read-data-python}. In a nutshell, this action reads the content of a data file and loads it into a dataframe. As we may see, it takes a list of parameters and the state as input. This signature is common to all the action implementations in \genplanx and allows the action itself to update the execution state that will be returned at the end of its execution.

\begin{lstlisting}[
  float=!htb,
  caption={read-data Python implementation.},
  label={lst:read-data-python},
  language=Python]
def read_data (parameters, state):
    # obtains the object names from the parameters of the action
    data_var = parameters[1]
    file_var = parameters[2]
    # gets the file path from the file dictionary in the execution state
    path = state[file_var]["value"]
    df = pd.read_csv(path)
    columns = df.columns.tolist()
    state[data_var] = {"type": "dataframe", "value": df, "columns": columns}
    # adds columns of dataframe as new objects to the execution state
    for col in columns:
        if col not in state:
            name = col.replace(" ", "-") + "_column"
            name = name.lower()
            state[name] = {"type": "column", "value": col}
    return state
\end{lstlisting}

This module will also periodically sense the real-world and to check whether the plan is going as expected or the execution/real-world has deviated from the planned behavior.
As opposed to other planning-execution architectures~\cite{icaps11-demo-pelea}, currently \genplanx only monitors the low-level state without translating it back to the high-level PDDL state.
More specifically, monitoring is done through boolean functions that check whether the execution of the action into the real-world yielded the expected effects.
Listing~\ref{lst:read-data-monitor} shows the python function that monitors the execution of the \texttt{read-data} action.

\begin{lstlisting}[
  float=!htb,
  caption={Function that monitors the success of the read-data action execution.},
  label={lst:read-data-monitor},
  language=Python]
def read_data_success (action, state):
    data_var = action[2]
    success = (isinstance(state[data_var]["value"], pd.DataFrame))
    return success
\end{lstlisting}

As we can see, this function checks whether the data read from the file is a dataframe instance or not. In case the execution of any action fails, \genplanx replans. Another reason for replanning consists on of an action execution adding a new goal to the execution state. As an example, a read-email action can read the contents of the email and generate as a new set of goals to achieve the intents expressed in the email.

\section{Examples}

\refstepcounter{genplanxexample}

\newcommand{\slideexamplesimple}{\arabic{genplanxexample}.1\xspace}
\newcommand{\slideexamplecomplex}{\arabic{genplanxexample}.2\xspace}

In this section, we present some examples that show the utility of \genplanx. For an office task related to annual report presentation, we present Example~\slideexamplesimple and demonstrate how \genplanx successfully solves the task. Then, in Example~\slideexamplecomplex, we highlight the importance of a classical planner for generating optimal plans, particularly in scenarios involving action costs and preconditions. In Section~\ref{sec:learning_example}, we present an example where the plan includes learning a prediction model and using it for new data. We also provide additional examples in the Appendix.

\subsection{Examples \slideexamplesimple and \slideexamplecomplex: Annual Report Presentation}
\begin{tcolorbox}[colback=blue!1!white, colframe=blue!5!black, title=User Request for Example \slideexamplesimple]
Read "annual-report.csv" and generate a barchart from "balance" against reference column "year". Create a slide with bar chart with title "Balance over years", and add it to a presentation. Save the presentation on file genplanx/graph.pptx.
\end{tcolorbox}

With the help of a classical planner, \genplanx is able to generate an optimal plan (as shown in Figure~\ref{fig:init_state_goal_and_plan_example1}). \genplanx solves this task end-to-end in several steps. \genplanx does integrated planning and execution for office tasks. It uses an LLM for generating the planning problem in PDDL. In the first step, \genplanx creates the PDDL planning task. In the second step, it uses a classical planner to solve the planning task and generate an optimal plan. In the third step, it executes the plan with online replanning whenever necessary. 

To create the PDDL planning task, \genplanx adds the intents (see Section~\ref{sec:prompting}) to the user request, and sends a query to the LLM. The LLM returns the dictionary with the variables and their initial values, the initial state and goals of the planning problem. The set of variables with their initial values are shown in Table~\ref{table:variables_example1}.

\begin{table}[hbt]
    \centering
    \rowcolors{2}{gray!10}{white}
    \begin{tabular}{>{\bfseries}l l p{8cm}} 
        \toprule
        \rowcolor{gray!20}
        \textbf{Variable} & \textbf{Type} & \textbf{Initial Value/Description} \\
        \midrule
        database1 & Database & `annual-report'\\
        presentation-file1 & PowerPoint file & ./genplanx/graph.pptx \\
        query1 & Query & df[`balance'] \\
        query2 & Query & df[`year'] \\
        column1 & Column & `balance' \\
        reference1 & Column & `year' \\
        title1 & Title & `Balance over years' \\
        presentation1 & Presentation & Empty; name: presentation-annual-reports \\
        slide1 & Slide & Empty; name: Slide 1 \\
        dataframe1 & Dataframe & Empty; dataframe to be populated\\
        filtered-dataframe & Dataframe & Empty; dataframe for filtered data \\
        bar-chart1 & Bar chart & Empty; name: 
    `Bar Chart 1'\\
        reference-dataframe & Dataframe & Empty; dataframe for reference data \\
        \bottomrule
    \end{tabular}
    \caption{Variables in Example \slideexamplesimple generated by the LLM  with their initial values.}
    \label{table:variables_example1}
\end{table}

\genplanx also uses the LLM and a compiler (see Section~\ref{sec:architecture}) to generate the initial state and goals of the planning problem. The initial state and the goals for this example, as generated, are shown in Figure~\ref{fig:init_state_and_goals_slide_example_simple}.

\begin{figure}
{\small
\smallskip

\begin{tabular}{@{}p{16cm}@{}}
\textbf{Initial State:} \\
\texttt{(in dataframe1 database1) }\\ 
\texttt{(available query1) (query-result dataframe1 query1 filtered-dataframe)}\\
\texttt{(reference reference1 reference-dataframe)} \\
\texttt{(available query2) (query-result dataframe1 query2 reference-dataframe)}\\
\texttt{(= (database-cost database1) 1)}\\
\texttt{(database-query-basic db1)}\\
\texttt{(= (total-cost) 0)}\\

\textbf{Goals:} \\
\texttt{(and} \\
\texttt{    (done-query query1) (done-query query2)} \\ 
\texttt{    (in-graph filtered-dataframe reference-dataframe bar-chart1) }\\
\texttt{    (in bar-chart1 slide1) (in title1 slide1)}\\
\texttt{    (in slide1 presentation1) }\\
\texttt{    (in presentation1 presentation-file1))} \\
\end{tabular}
\smallskip}
\caption{The initial state and goals for Example~\slideexamplesimple.}
\label{fig:init_state_and_goals_slide_example_simple}
\end{figure}

\begin{figure}[!h]
    \centering
    \setlength{\fboxsep}{0pt} 
    \setlength{\fboxrule}{1pt} 
    \fbox{\includegraphics[width=0.75\linewidth]{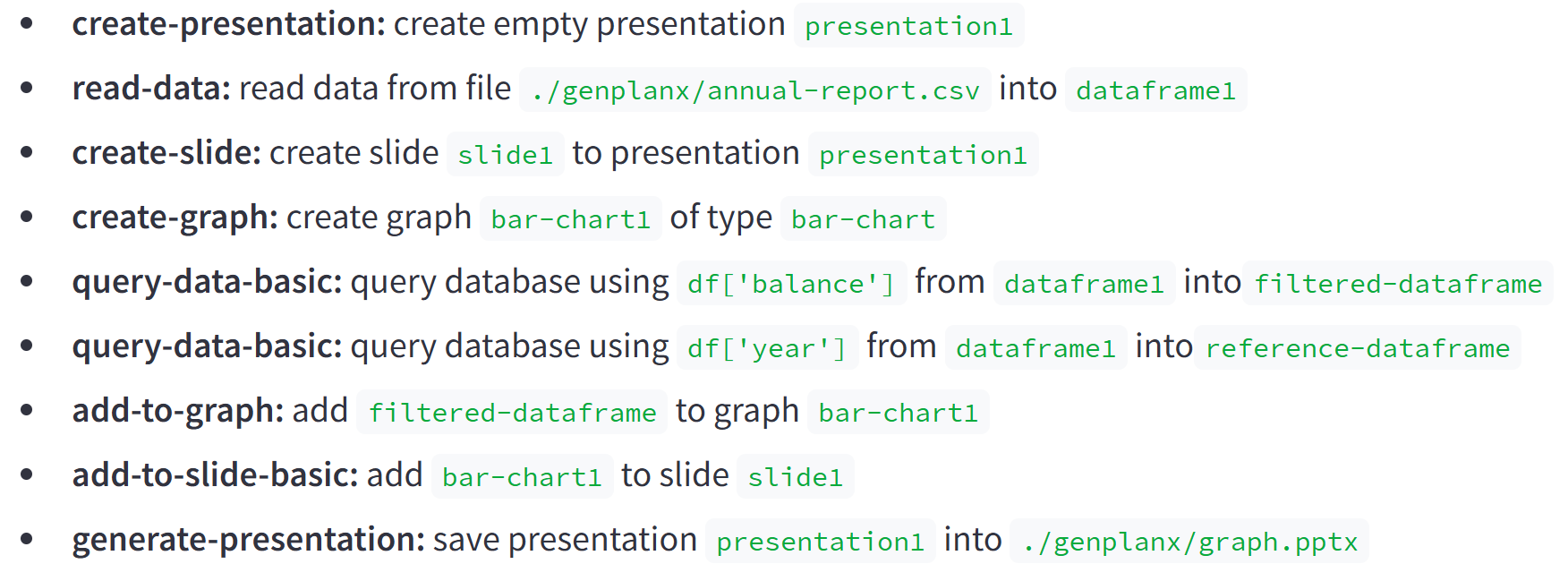}}
    \caption{Step by step plan generated and executed by \genplanx for Example \slideexamplesimple.}
    \label{fig:step_by_step_example1}
\end{figure}

\begin{figure}[hbt]
    \centering
    \setlength{\fboxsep}{0pt} 
    \setlength{\fboxrule}{1pt} 

      \fbox{\includegraphics[width=0.4\linewidth]{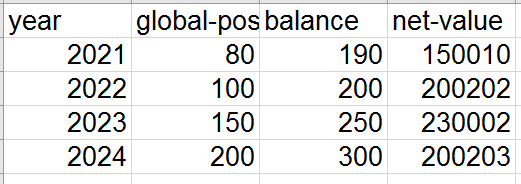}}\hspace{0.5cm}
    \fbox{\includegraphics[width=0.55\linewidth]{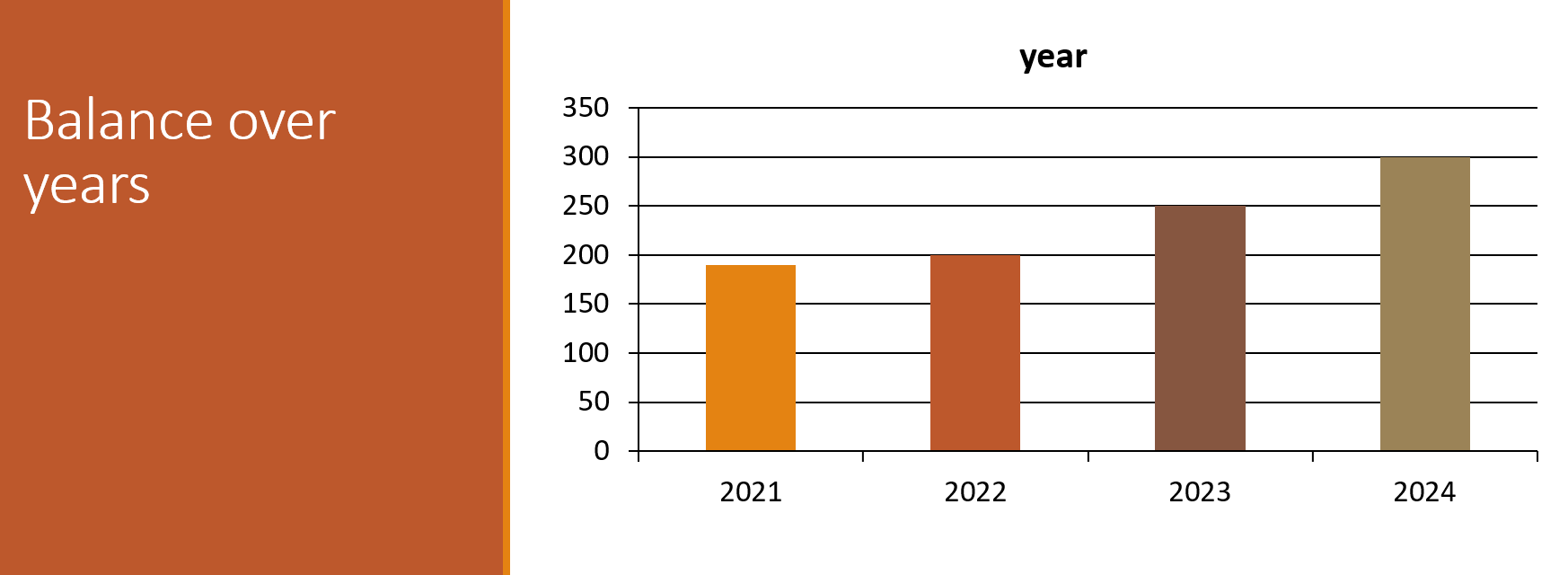}}
    \caption{The data file (left) and the final result, Powerpoint Slide (right) created for Example \slideexamplesimple and \slideexamplecomplex.}
    \label{fig:output_slide_example}
\end{figure}

The plan for this example is shown in  Figure~\ref{fig:init_state_goal_and_plan_example1}. Finally, \genplanx executes this plan step by step (Figure ~\ref{fig:step_by_step_example1}) and creates the corresponding Powerpoint file with the correct information (Figure~\ref{fig:output_slide_example}).

\begin{figure}[!h]
\centering
{\small
\begin{tikzpicture}[node distance=1.3cm, auto]

    \tikzstyle{state} = [rectangle, draw, text width=15cm, text centered, minimum height=2em]
    \tikzstyle{action} = [rectangle, draw, fill=blue!5, text width=10cm, text centered, minimum height=2em]
    \tikzstyle{goal} = [rectangle, draw, fill=green!20, text width=15cm, text centered, minimum height=2em]

    \node[state] (initial) {Initial State: \texttt{(in dataframe1 database1) (available query1) (query-result dataframe1 query1 filtered-dataframe) (reference reference1 reference-dataframe) (available query2) (query-result dataframe1 query2 reference-dataframe) (= (database-cost database1) 1) (database-query-basic db1) (= (total-cost) 0)}};

    \node[action, below of=initial] (action1) {\texttt{create-graph(ai, bar-chart1)}};
    \node[action, below of=action1] (action2) {\texttt{read-data(ai, dataframe1, database1)}};
    \node[action, below of=action2] (action3) {\texttt{create-presentation(ai, presentation1)}};
    \node[action, below of=action3] (action4) {\texttt{create-slide(ai, slide1, presentation1, title1)}};
    \node[action, below of=action4] (action5) {\texttt{query-data-basic(ai, query1, dataframe1, filtered-dataframe, database1)}};
    \node[action, below of=action5] (action6) {\texttt{query-data-basic(ai, query2, dataframe1, reference-dataframe, database1)}};
    \node[action, below of=action6] (action7) {\texttt{add-to-graph(ai, filtered-dataframe, bar-chart1, reference1, reference-dataframe)}};
    \node[action, below of=action7] (action8) {\texttt{add-to-slide-basic(ai, bar-chart1, slide1, presentation1)}};
    \node[action, below of=action8] (action9) {\texttt{generate-presentation(ai, presentation1, presentation-file1)}};

    \node[goal, below of=action9] (goal) {Goals: \texttt{(and (done-query query1) (done-query query2) (in-graph filtered-dataframe reference-dataframe bar-chart1) (in bar-chart1 slide1) (in title1 slide1) (in slide1 presentation1) (in presentation1 presentation-file1))}};

    \draw[->] (initial) -- (action1);
    \draw[->] (action1) -- (action2);
    \draw[->] (action2) -- (action3);
    \draw[->] (action3) -- (action4);
    \draw[->] (action4) -- (action5);
    \draw[->] (action5) -- (action6);
    \draw[->] (action6) -- (action7);
    \draw[->] (action7) -- (action8);
    \draw[->] (action8) -- (action9);
    \draw[->] (action9) -- (goal);

\end{tikzpicture}
}
\caption{Initial state, goals and plan for Example \slideexamplesimple in PDDL. The initial state and the goals are generated by the LLM and a compiler. The plan is generated using a classical planner.}
\label{fig:init_state_goal_and_plan_example1}
\end{figure}

Next, we present a variation of this example where there are multiple possible plans, and show that \genplanx generates and executes the optimal plan (the plan with the lowest cost). 

\begin{tcolorbox}[colback=blue!1!white, colframe=blue!5!black, title=User Request for Example \slideexamplecomplex (Two choices of database to read data from)]
Read data and generate a barchart from `balance' against reference column `year’. 
The available databases to read from are db1 with cost of reading 1 and db2 with cost of reading 3. 
db1 supports basic query. db2 supports optimized query. 
Create a slide with bar chart with title `Balance over years', and add it to a presentation. Save the presentation on file genplanx/graph.pptx.
Create a slide with bar chart with title `Balance over years', and add it to a presentation.
Save the presentation on file ./genplanx/graph.pptx
\end{tcolorbox}

From the user request of Example \slideexamplecomplex, we can observe that \genplanx first needs to read the data from a database. Consider the situation where the data can be accessed from two databases, \texttt{db1} and \texttt{db2}, which may differ in access protocols and design. 
 The initial state and the goals are in Figure~\ref{fig:init_state_and_goal_slide_example_complex}.
{\small
\smallskip
\begin{figure}
\begin{tabular} {@{}p{16cm}@{}}
\textbf{Initial State:} \\
\texttt{(in dataframe1 db1) in dataframe1 db2)}\\ 
\texttt{(available query1) (query-result dataframe1 query1 filtered-dataframe)}\\
\texttt{(reference reference1 reference-dataframe)} \\
\texttt{(available query2) (query-result dataframe1 query2 reference-dataframe)}\\
\texttt{(database-query-optimized db2) (database-query-basic db1)}\\
\texttt{(= (database-cost database1) 1) (= (database-cost database2) 2)}\\
\texttt{(= (total-cost) 0)}\\
\end{tabular}
\begin{tabular} {@{}p{16cm}@{}}
\textbf{Goals:}\\
\texttt{(and} \\
\texttt{    (done-query query1) (done-query query2)} \\ 
\texttt{    (in-graph filtered-dataframe reference-dataframe bar-chart1) }\\
\texttt{    (in bar-chart1 slide1) (in title1 slide1)}\\
\texttt{    (in slide1 presentation1) }\\
\texttt{    (in presentation1 presentation-file1))} \\
\end{tabular}
\smallskip
\caption{Initial state and goals for Example\slideexamplecomplex}
\label{fig:init_state_and_goal_slide_example_complex}
\end{figure}
}

The read step has two alternative choices for action parameter, \texttt{read-data(db1)} with cost 1 and \texttt{read-data(db2)} with cost 2. 
The choice made at the read step has an effect on the query step. \texttt{db1} only support basic query (action \texttt{query-data-basic}) with cost 5, and \texttt{db2} supports optimized query (action \texttt{query-data-optimized}) with cost 2.
The possible choices for reading and querying data for this task are in Table~\ref{table:plan_choices}

\begin{center}
\begin{table}[hbt]
\centering
\begin{tabular}{lll}
\toprule
\textbf{Read Action} & \textbf{Query Action} & \textbf{Total Cost} \\
\midrule
\texttt{read-data(db1)} & \texttt{query-data-basic} & 1 + 5 = 6 \\
\texttt{read-data(db1)} & \texttt{query-data-optimized} & Invalid \\
\rowcolor{green} \texttt{read-data(db2)} & \texttt{query-data-optimized} & 2 + 2 = 4 \\
\texttt{read-data(db2)} & \texttt{query-data-basic} & Invalid \\
\bottomrule
\end{tabular}
\caption{Possible action choices for reading and querying data for the planning task in Example \slideexamplecomplex.}
\label{table:plan_choices}
\end{table}
\end{center}

\begin{figure}[!h]
\centering
{\small
\begin{tikzpicture}[node distance=1.4cm, auto]

    \tikzstyle{state} = [rectangle, draw, text width=15cm, text centered, minimum height=2em]
    \tikzstyle{action} = [rectangle, draw, fill=blue!5, text width=15cm, text centered, minimum height=2em]
    \tikzstyle{goal} = [rectangle, draw, fill=green!20, text width=15cm, text centered, minimum height=2em]

    \node[state] (initial) {Initial State: \texttt{(in dataframe1 db1) (in dataframe1 db2) (available query1) (query-result dataframe1 query1 filtered-dataframe) (reference reference1 reference-dataframe) (available query2) (query-result dataframe1 query2 reference-dataframe) (database-query-optimized db2) (database-query-basic db1) (= (database-cost database1) 1) (= (database-cost database2) 2) (= (total-cost) 0)}};

    \node[action, below of=initial] (action1) {\texttt{create-graph(ai, bar-chart1)}};
    \node[action, below of=action1] (action2) {\texttt{read-data(ai, dataframe1, database2)}};
    \node[action, below of=action2] (action3) {\texttt{create-presentation(ai, presentation1)}};
    \node[action, below of=action3] (action4) {\texttt{create-slide(ai, slide1, presentation1, title1)}};
    \node[action, below of=action4] (action5) {\texttt{query-data-optimized(ai, query1, dataframe1, filtered-dataframe, database2)}};
    \node[action, below of=action5] (action6) {\texttt{query-data-optimized(ai, query2, dataframe1, reference-dataframe, database2)}};
    \node[action, below of=action6] (action7) {\texttt{add-to-graph(ai, filtered-dataframe, bar-chart1, reference1, reference-dataframe)}};
    \node[action, below of=action7] (action8) {\texttt{add-to-slide-basic(ai, bar-chart1, slide1, presentation1)}};
    \node[action, below of=action8] (action9) {\texttt{generate-presentation(ai, presentation1, presentation-file1)}};

    \node[goal, below of=action9] (goal) {Goals: \texttt{(and (done-query query1) (done-query query2) (in-graph filtered-dataframe reference-dataframe bar-chart1) (in bar-chart1 slide1) (in title1 slide1) (in slide1 presentation1) (in presentation1 presentation-file1))}};

    \draw[->] (initial) -- (action1);
    \draw[->] (action1) -- (action2);
    \draw[->] (action2) -- (action3);
    \draw[->] (action3) -- (action4);
    \draw[->] (action4) -- (action5);
    \draw[->] (action5) -- (action6);
    \draw[->] (action6) -- (action7);
    \draw[->] (action7) -- (action8);
    \draw[->] (action8) -- (action9);
    \draw[->] (action9) -- (goal);

\end{tikzpicture}
}

\caption{The initial state, goals and plan for Example \slideexamplecomplex in PDDL. The initial state and the goals are generated by the LLM and a compiler of \genplanx. The plan is generated using a classical planner.}
\label{fig:init_state_goal_and_plan_example1.2}
\end{figure}
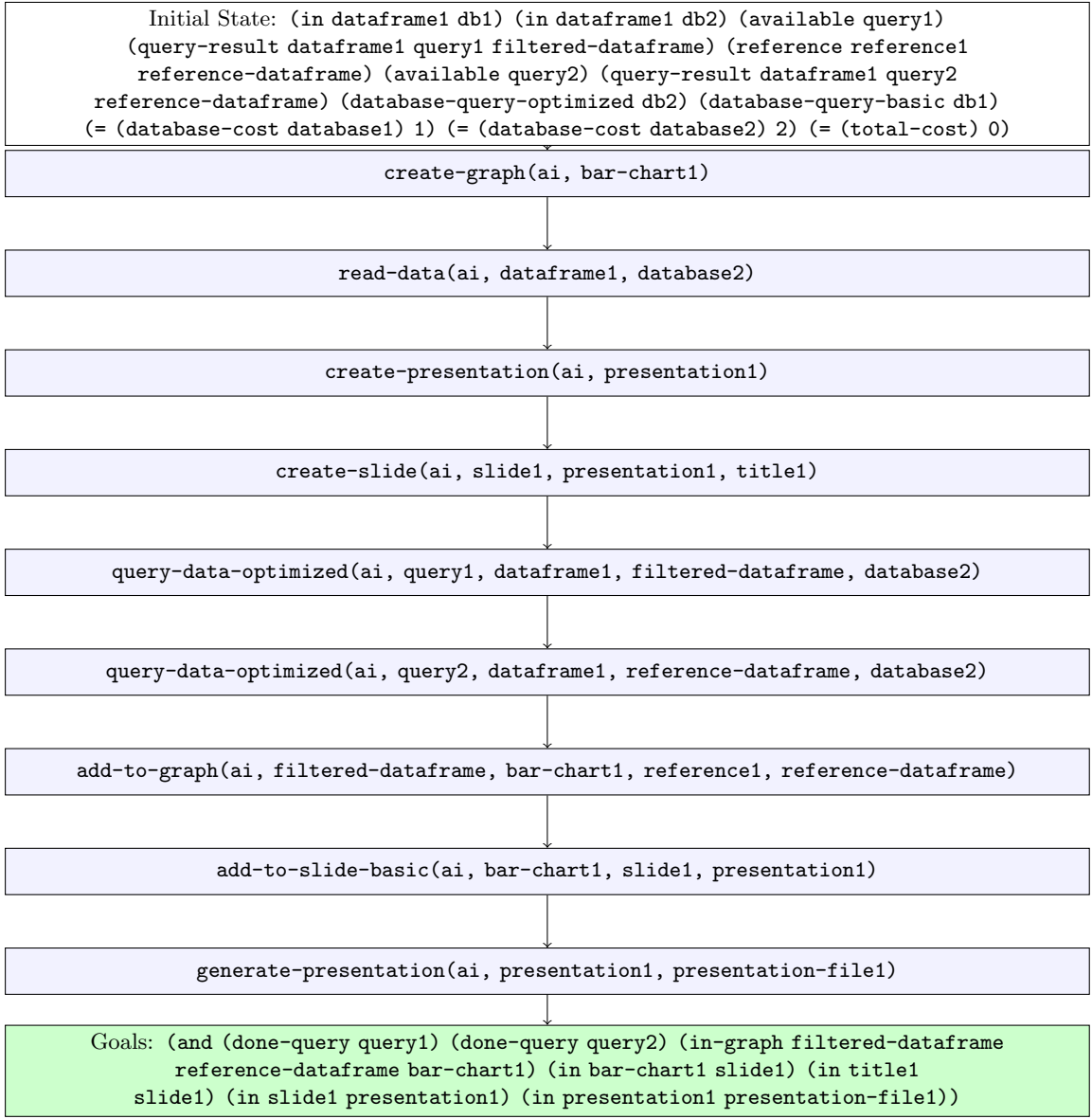

We can observe that the optimal plan should include \texttt{read-data(db2)} and \texttt{query-data-optimized}.
With the help of a classical planner, \genplanx is able to generate an optimal plan (as shown in Figure~\ref{fig:init_state_goal_and_plan_example1.2}). 
When we asked the LLM, GPT-4O, to generate a plan for this task, it was unable to generate the optimal plan. The details are in Appendix~\ref{sec:llm_for_planning}. In the appendix, we also show that even the o3-mini reasoning model does not generate correct or optimal plan consistently without additional hints (the hints are elaborated in the appendix).

\begin{figure}[hbt]
    \centering
    \setlength{\fboxsep}{0pt} 
    \setlength{\fboxrule}{1pt} 
    \fbox{\includegraphics[width=0.75\linewidth]{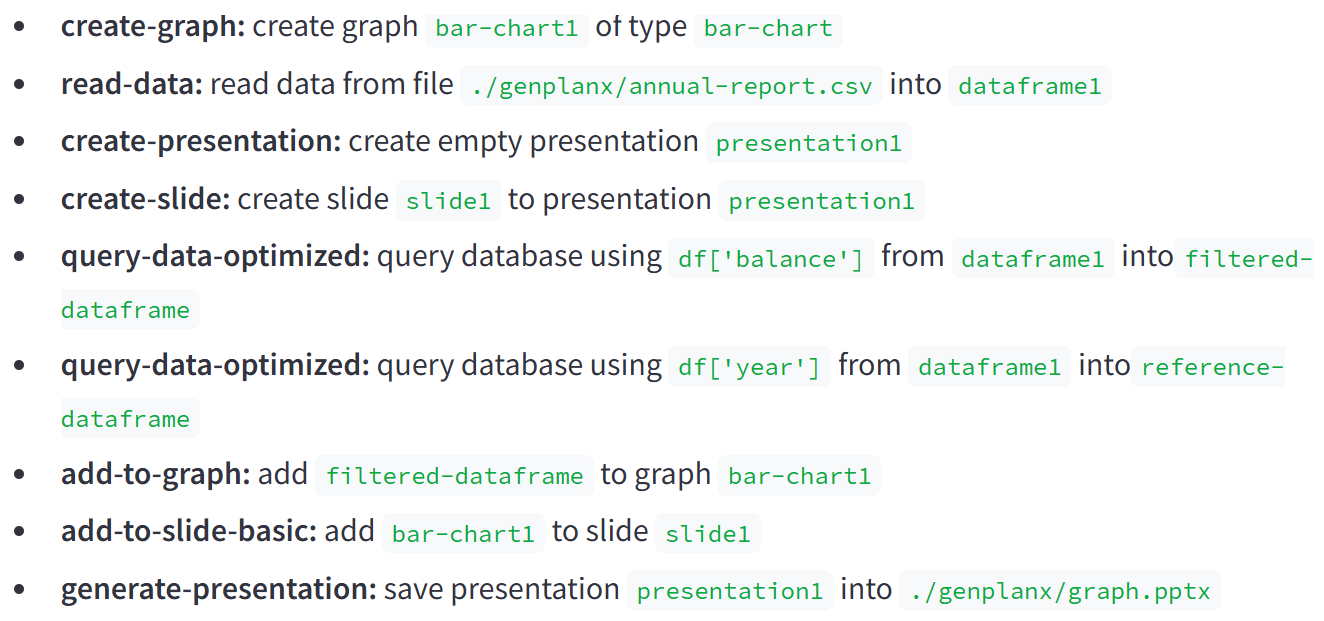}}
    \caption{Step by step plan generated and executed by \genplanx for Example \slideexamplecomplex}
    \end{figure}


\refstepcounter{genplanxexample}
\edef\learningexample{\arabic{genplanxexample}}

\subsection{Example \learningexample: Investor Prediction with Decision Tree}
\label{sec:learning_example}
\begin{tcolorbox}[colback=blue!1!white, colframe=blue!5!black, title=User Request for Example \learningexample]
Read ./genplanx/investment\_data.csv and ./genplanx/investment\_pred.csv. The available database for both files is db with cost of reading, 1.
 Learn the WillPurchase column with the DecisionTreeClassifier algorithm.
The name of the model is investor. Use the trained model to predict the WillPurchase column for the data in the file ./genplanx/investment\_pred.csv
\end{tcolorbox}

Now, we present an example where the plan includes learning a prediction model and using it for new data. From the user request, we can observe that \genplanx first needs to read the data from a file `genplanx/investment\_data.csv'. This data should be used to train a decision tree classifier for the `WillPurchase' column, and the model should be called ``investor''. 
The set of variables with their initial values can be found in Table~\ref{table:investment_variables}.
The initial state and the goals are generated by the LLM and a compiler in Figure~\ref{fig:init_state_and_goal_learning_example}.

\begin{figure}
{\small
\smallskip

\begin{tabular}{@{}p{16cm}@{}}
\textbf{Initial State:} \\
\texttt{(in dataframe1 data-file1)}\\
\texttt{(in dataframe2 data-file2)}\\
\texttt{(= (database-cost database1) 1)}\\
\texttt{(= (total-cost) 0)}\\

\textbf{Goals:} \\
\texttt{(and} \\
\texttt{ (learned-model model1 column1)}\\
\texttt{ (used-for-training dataframe1)}\\
\texttt{ (done-prediction model1 dataframe2 column1 dataframe3)}\\
\texttt{ (available dataframe3))}\\
\end{tabular}
\smallskip
}
\caption{The initial state and goals in Example \learningexample.}
\label{fig:init_state_and_goal_learning_example}
\end{figure}

\begin{figure}[hbt]
    \centering
    \fbox{\includegraphics[width=\linewidth]{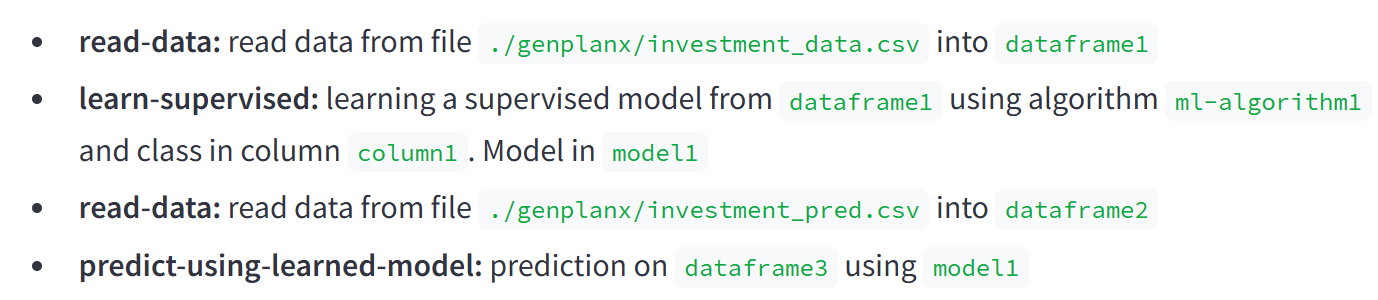}}
    \caption{Plan generation and execution for User Request 2.}
    \label{fig:learning_example}
\end{figure}

\begin{figure}[hbt]
    \centering
    \setlength{\fboxsep}{0pt} 
    \setlength{\fboxrule}{1pt} 
  \fbox{\includegraphics[width=0.75\linewidth]{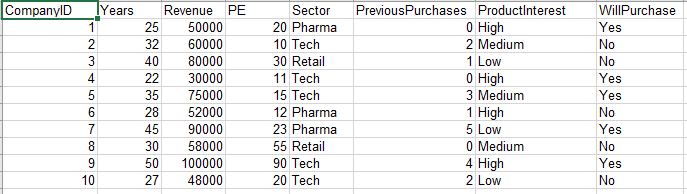}}
    \fbox{\includegraphics[width=0.75\linewidth]{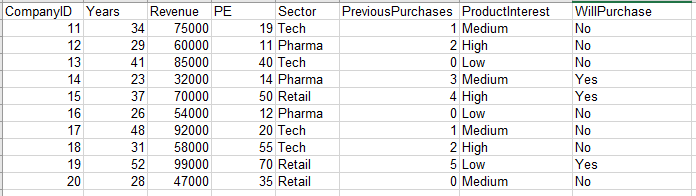}}
    \caption{The input file (top) and the final result, prediction of the `WillPurchase' column using learned model for the (bottom) Example \learningexample.}
    \label{fig:output_example2}
\end{figure}

\genplanx needs to read the data from a file `genplanx/investment\_pred.csv' and use the model ``investor'' to predict the values of `WillPurchase' for the records in `investment\_pred.csv'. The plan generated by \genplanx is shown in Figure~\ref{fig:learning_example_plan} and the steps of executing the plan is shown in Figure~\ref{fig:learning_example}. First, \genplanx reads the training data file, then it learns the desired model on the given target column. Finally, \genplanx reads the file containing the records to predict, applies the model to the records, and stores the predictions in a new dataframe (Figure~\ref{fig:output_example2}).

\begin{table}[hbt]
    \centering
    \rowcolors{2}{gray!10}{white}
    \begin{tabular}{>{\bfseries}l l p{8cm}} 
        \toprule
        \rowcolor{gray!20}
        \textbf{Variable} & \textbf{Type} & \textbf{Initial Value/Description} \\
        \midrule
        data-file1      & Data file    & ../genplanx/investment\_data.csv \\
        data-file2      & Data file    & ../genplanx/investment\_pred.csv \\
        database1       & Database     & db \\
        dataframe1      & Dataframe    & Empty; dataframe to be populated with data-file1\\
        dataframe2      & Dataframe    & Empty; dataframe to be populated with data-file2\\
        dataframe3      & Dataframe    & Empty; dataframe to be populated with model predictions \\
        column1         & Column       & `WillPurchase' \\
        model1          & Model        & ``investor'', algorithm: ml-algorithm1; decision tree model\\
        ml-algorithm1   & ML algorithm & DecisionTreeClassifier \\
        \bottomrule
    \end{tabular}
    \caption{Variables in the investment data configuration with their initial values.}
    \label{table:investment_variables}
\end{table}

\begin{figure}[hbt]
    \centering
    \begin{tikzpicture}[node distance=1.5cm, auto, thick, 
        state/.style={rectangle, draw, text width=12cm, text centered, minimum height=1cm, font=\ttfamily},
        plan/.style={rectangle, draw, text width=10cm, text centered, minimum height=1cm, fill=blue!10, font=\ttfamily},
        goal/.style={rectangle, draw, text width=12cm, text centered, minimum height=1cm, fill=green!20, font=\ttfamily},
        every node/.style={font=\ttfamily}]

        \node[state] (init) {Initial State: (in dataframe1 data-file1) (in dataframe2 data-file2) (= (database-cost database1) 1)};

        \node[plan, below of=init] (step1) {read-data(ai, dataframe1, data-file1, database1)};
        \node[plan, below of=step1] (step2) {learn-supervised(ai, dataframe1, ml-algorithm1, column1, model1)};
        \node[plan, below of=step2] (step3) {read-data(ai, dataframe2, data-file2, database1)};
        \node[plan, below of=step3] (step4) {predict-using-learned-model(ai, dataframe2, column1, model1, dataframe3, ml-algorithm1)};

        \node[goal, below of=step4] (goal) {Goals: (and (learned-model model1 column1) (used-for-training dataframe1) (done-prediction model1 dataframe2 column1 dataframe3) (available dataframe3))};

        \draw[->] (init) -- (step1);
        \draw[->] (step1) -- (step2);
        \draw[->] (step2) -- (step3);
        \draw[->] (step3) -- (step4);
        \draw[->] (step4) -- (goal);

    \end{tikzpicture}
    \caption{The initial state, goals and plan for Example \learningexample in PDDL. The initial state and the goals are
generated by the LLM. The plan is generated using a classical planner.}
\label{learning_example_plan}
    \label{fig:learning_example_plan}
\end{figure}

\section{Conclusions and Future work}
In this paper we have introduced \genplanx. It can understand user requests in natural language, generate plans to address those requests, execute the plans in a real office environment, and monitor the execution. We have contributed with a new domain that implements common office-related actions, the implementation of those actions, as well as an architecture that integrates LLMs and AI classical planning.

In future work we would like to improve \genplanx in two main fronts.
First, we would like to expand the set of actions considered in order to cover a wider ranger of tasks.
This process could be either manual as we are currently doing; or automated, by learning action models from observations~\cite{AinetoCO19,morales2024learning,gragera2023planning}.
Second, we would like to provide \genplanx with goal reasoning~\cite{DBLP:conf/iccbr/Munoz-AvilaJAC10} capabilities.
In particular, we would like to let \genplanx automatically generate new goals upon replanning~\cite{ecai23-replanning}, when monitoring detects opportunities upon changes in the environment~\cite{icaps21-opportunities}, by analyzing the structure of the goals~\cite{icaps24-symbolic-centroids}, or by predicting the appearance of new goals~\cite{aicomm18-learning,aicomm18-anticipatory}.

\section*{Disclaimer}
This paper was prepared for informational purposes by
the Artificial Intelligence Research group of JPMorgan Chase \& Co. and its affiliates (``JP Morgan''),
and is not a product of the Research Department of JP Morgan.
JP Morgan makes no representation and warranty whatsoever and disclaims all liability,
for the completeness, accuracy or reliability of the information contained herein.
This document is not intended as investment research or investment advice, or a recommendation,
offer or solicitation for the purchase or sale of any security, financial instrument, financial product or service,
or to be used in any way for evaluating the merits of participating in any transaction,
and shall not constitute a solicitation under any jurisdiction or to any person,
if such solicitation under such jurisdiction or to such person would be unlawful.

\bibliographystyle{plain}
\bibliography{general, daniel, references}

\section{Appendix 1}

\subsection{Complete prompt used to call the LLM}
\label{app_example_prompt}

We include here an example of a full prompt that is used as input to the LLM.

\begin{lstlisting}
    
You are working in an office environment. You have to provide responses to queries from employees or clients related to sales. Information is present on several systems of records (SOR). The schemas (columns) of those systems and their values are the following:
Schema for SOR 1: [list of columns], SOR 2: [list of columns]
Pay attention to the upper or lower case of the fields in the provided schema when creating queries to the databases. Given a request, choose the appropriate SOR, and identify a set of intents on that request. Then, return a Python dictionary that contains information on all intents. You should not define a function or provide python code, but return the dictionary as your output. Do not use external tools. The keys of the dictionary are the elements of the task (entities), the 'init_state' and the 'goals'. All keys have to be in lower case. 'init_state' and 'goals' are mandatory. The values of the dictionary keys are definition dictionaries. A definition dictionary is a python dictionary, where the keys are 'type', 'value', and some other element specific keys, such as 'to', 'body', or 'subject' for emails. The types of the task are: pie-chart, bar-chart, histogram, column, value-counts, count, value, input-email, output-email, human-agent, ai-agent, excel-file, csv-file, dataframe, text, graph, title, api, data-file, pdf-file, word-file, text-file, powerpoint-file, row, file, email, data, model, data-contents, response, query, chat-history, ml-algorithm, presentation, appointments, appointments-item, slide, contents, data-type, agent, language, session, object. The entities' values should be extracted from the intents on the request.  You cannot use as keys of dictionary the names of types and you cannot use repeated keys. The value of 'init_state' is a string whose contents is a state, where a state is a sequence of literals separated by spaces. The init_state represents what is known to be true at the beginning. Each literal is a tuple whose elements are separated by spaces. The first element of the literals is a predicate from the following list where each element is of the form (<predicate-name> <parameters>):
	(in ?c - contents ?c1 - contents)
	(in-data ?dt - data-type ?d - contents)
	(in-graph ?c - contents ?c1 - contents ?c3 - graph)
	(available ?o - object)
	(used ?c - contents)
	(data-type-contents ?dt - data-type ?dc - data-contents)
	(web-search-result ?q - text ?r - text)
	(query-result ?d - dataframe ?q - query ?d1 - data)
	(answer-llm ?q - text ?r - text)
	(done-merge)
	(reference ?c - column ?d - dataframe)
	(done-query ?q - query)
	(done-question ?q - text)
	(modified ?d - dataframe ?c - contents ?co - column ?t - text ?d1 - dataframe)
	(added-value ?d - dataframe ?c - contents ?co - column ?t - text ?d1 - dataframe)
	(merged ?d - dataframe ?d1 - dataframe ?d2 - dataframe)
	(deleted ?d - dataframe ?d1 - dataframe ?d2 - dataframe)
	(sent-contents ?rc - contents ?r - response)
	(sent ?r - response)
	(sent-email ?e - output-email ?c - contents)
	(read-email-contents ?c - contents ?f - file ?e - input-email)
	(replied-email ?e - input-email)
	(email-read ?e - input-email)
	(email-parsed ?e - input-email)
	(notified ?e - output-email ?c - contents)
	(explained ?c - contents ?c1 - contents)
	(translated ?c - contents ?l - language ?c1 - contents)
	(summarized ?c - contents ?c1 - contents)
	(search-result ?t - text ?q - text ?t1 - text)
	(info-on ?d - data-contents ?g - graph ?p - presentation)
	(appointments-read ?s - appointments)
	(appointments-contents ?s - appointments ?d - dataframe)
	(learned-model ?m - model ?d - dataframe ?c - column)
The parameters are defined as: <variable> - <type>. The value of 'goals' is the list of intents, also represented as a state (list of literals). Take the types into account when defining the literals in the states (init_state and goal). For example, if you want to express that a file F contains a dataframe D, add to the init_state the literal '(in D F)'. You cannot use elements of other types as arguments of the corresponding predicate. Make sure all parameters of all literals that appear in the 'init_state' and 'goals' have an entry in the output dictionary. Everything that is true at start should be in the 'init_state'. Everything that you would like to be true at the end should be specified in the 'goals'. If you find more than one intent in the request, merge the dictionaries into a single dictionary. In order to merge the dictionaries, add all entities found. Also, the merged 'init_state' will be the list of all literals in all the intents' 'init_state'. Likewise, the merged 'goals' will be the list of all literals on all intents' 'goals'. Do not include literals from the goals in the initial state. This is important: only use one chat response. Make sure you format the output properly and take into account all the previous constraints. When creating queries to databases please take into account semantics. As an example, if the request asks about not matched transactions, check for all semantically equivalent values, as unmatched. Do not return Output: in the output.
I will give you now several examples with their corresponding output as '
Intent: <intent>
Output:
<dictionary>'.
Examples:

Intent: Summarize
Output:
{'text1': {"type": "text", "value": "matched"},  'text2': {'type': 'text', 'value': 'text2'}, 'init_state': {'type': 'state', 'value': ''}, 'goals': {'type': 'state', 'value': '(and (summarized text1 text2))'}}

Intent: Explain
Output:
{'text1': {"type": "text", "value": "matched"},  'text2': {'type': 'text', 'value': 'text2'}, 'init_state': {'type': 'state', 'value': '(available text1)'}, 'goals': {'type': 'state', 'value': '(and (explained text1 text2))'}}

Intent: Unknown Intent/Anything else/Something unrelated to the above intents
Output:
{"chat-response": {"type": "response", "value": "Apologies, I'm not able to help with that. Try another question!"}, "init_state": {"type": "state", "value": "(available chat-response)"}, "goals": {"type": "state", "value": "(and (sent chat-response))"}}

Intent:

\end{lstlisting}

\subsection{Additional Examples}

We will show now another example of the use of \genplanx. The input request is shown in Figure~\ref{fig:ppt_example2}. From the user request, we can observe that \genplanx first needs to read the data from the user's appointments. Then it should save this data into a CSV file named `apps.csv'.


\refstepcounter{genplanxexample}
\edef\appointmentexample{A\arabic{genplanxexample}}
\begin{tcolorbox}[colback=blue!1!white, colframe=blue!5!black, title=User Request for Example \appointmentexample: Part 1]
Read my appointments and save the result in apps.csv.
\end{tcolorbox}

The set of variables with their initial values are shown in Table~\ref{table:appointments_variables}.
The initial state and the goals are generated by the LLM and a compiler as in Table~\ref{table:init_and_goal_state_appointment_example}.

\begin{table}[!h]
{\small
\smallskip

\begin{tabular}{@{}p{16cm}@{}}
\textbf{Initial State:} \\
\texttt{(in appointments1 apps-file)}\\
\texttt{(available dataframe1)}\\
\texttt{(in dataframe1 apps-file)}\\

\textbf{Goals:} \\
\texttt{(and} \\
\texttt{ (appointments-contents appointments1 dataframe1)}\\
\texttt{ (available dataframe1)}\\
\texttt{ (in dataframe1 apps-file))}\\
\end{tabular}
\smallskip
}
\caption{Initial state and goals of Example \appointmentexample.}
\label{table:init_and_goal_state_appointment_example}
\end{table}

\begin{table}[hbt]
    \centering
    \rowcolors{2}{gray!10}{white}
    \begin{tabular}{>{\bfseries}l l p{8cm}} 
        \toprule
        \rowcolor{gray!20}
        \textbf{Variable} & \textbf{Type} & \textbf{Initial Value/Description} \\
        \midrule
        appointments1       & Appointments  & Empty; appointments data \\
        apps-file           & CSV File      & 'apps.csv'; csv file \\
        dataframe1          & Dataframe     & Empty; dataframe to populate with appointments data \\
        \bottomrule
    \end{tabular}
    \caption{Variables in the appointments data configuration with their initial values for Example~\appointmentexample.}
    \label{table:appointments_variables}
\end{table}

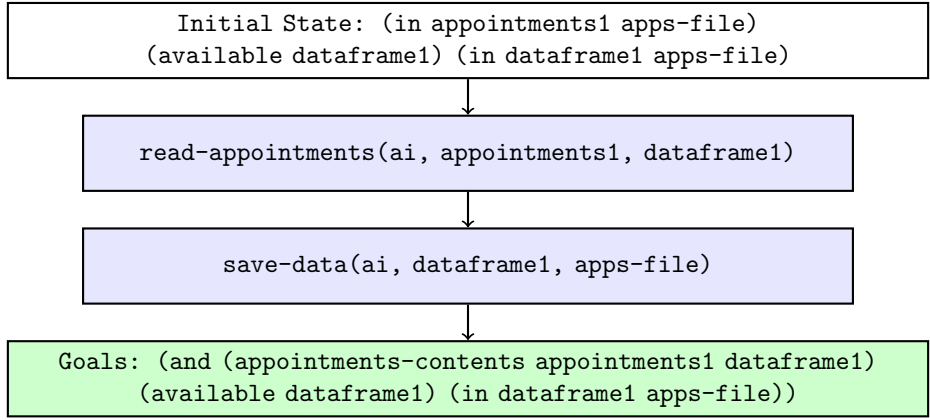
\begin{figure}[hbt]
    \centering
    \begin{tikzpicture}[node distance=1.5cm, auto, thick, 
        state/.style={rectangle, draw, text width=12cm, text centered, minimum height=1cm, font=\ttfamily},
        plan/.style={rectangle, draw, text width=10cm, text centered, minimum height=1cm, fill=blue!10, font=\ttfamily},
        goal/.style={rectangle, draw, text width=12cm, text centered, minimum height=1cm, fill=green!20, font=\ttfamily},
        every node/.style={font=\ttfamily}]

        \node[state] (init) {Initial State: (in appointments1 apps-file) (available dataframe1) (in dataframe1 apps-file)};

        \node[plan, below of=init] (step1) {read-appointments(ai, appointments1, dataframe1)};
        \node[plan, below of=step1] (step2) {save-data(ai, dataframe1, apps-file)};

        \node[goal, below of=step2] (goal) {Goals: (and (appointments-contents appointments1 dataframe1) (available dataframe1) (in dataframe1 apps-file))};

        \draw[->] (init) -- (step1);
        \draw[->] (step1) -- (step2);
        \draw[->] (step2) -- (goal);

    \end{tikzpicture}
    \caption{The initial state, goals, and plan for the appointments data analysis. The initial state and the goals are generated by the LLM. The plan is generated using a classical planner.}
    \label{fig:appointments_example_plan}
\end{figure}

\begin{figure}[hbt]
    \centering
    \setlength{\fboxsep}{0pt} 
    \setlength{\fboxrule}{1pt} 
    \fbox{\includegraphics[width=0.6\linewidth]{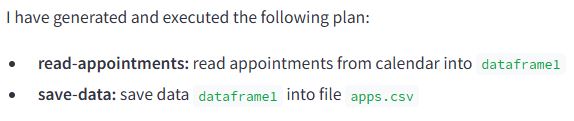}}
    \caption{Step by step plan generated and executed by \genplanx for Example~\appointmentexample}
    \label{fig:step_by_step_example3}
\end{figure}

\begin{figure}[hbt]
    \centering
    \fbox{\includegraphics[width=\linewidth]{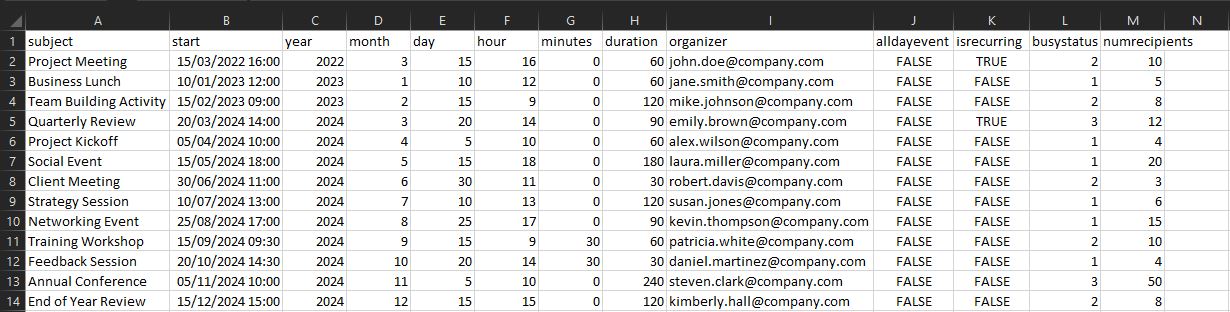}}
    \caption{Appoinments file created by \genplanx}
    \label{fig:ppt_example2}
\end{figure}

\begin{figure}[hbt]
    \centering
    \includegraphics[width=0.7\linewidth]{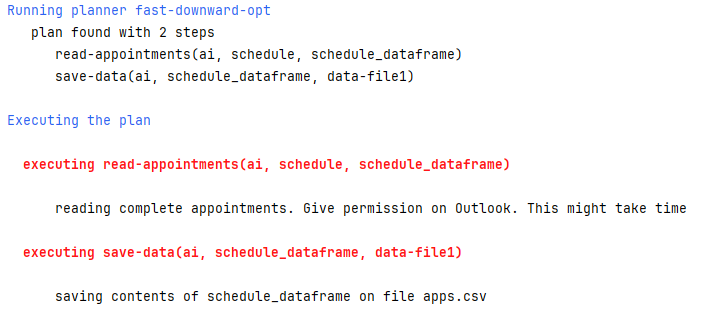}
    \caption{Connecting to Outlook and saving the appointments.}
    \label{fig:enter-label}
\end{figure}

\begin{tcolorbox}[colback=blue!1!white, colframe=blue!5!black, title=User Request for Appointment Filtering]
Open `apps.csv' and filter the appointments in 2024. Of these appointments count how many are at an hour between 8 and 11 and the number of recurring appointments (isrecurring).
\end{tcolorbox}
From the user request, we can observe that \genplanx first needs to read the data from the file `apps.csv'. Then it should filter the appointments for the year 2024, count the number of appointments within the specified hour range, and count the number of recurring appointments.

The set of variables with their initial values are in Table~\ref{table:appointments_filtering_variables}.
The initial state and the goals are generated by the LLM and a compiler as follows:

{\small
\smallskip

\begin{tabular}{@{}p{16cm}@{}}
\textbf{Initial State:} \\
\texttt{(in dataframe1 data-file1)}\\
\texttt{(available query1)}\\
\texttt{(query-result dataframe1 query1 filtered-2024-dataframe)}\\
\texttt{(available query2)}\\
\texttt{(query-result filtered-2024-dataframe query2 filtered-time-dataframe)}\\
\texttt{(available query3)}\\
\texttt{(query-result filtered-2024-dataframe query3 filtered-recurring-dataframe)}\\
\texttt{(in-data hour-counts filtered-time-dataframe)}\\
\texttt{(data-type-contents hour-counts hour-data-counts)}\\
\texttt{(in-data recurring-counts filtered-recurring-dataframe)}\\
\texttt{(data-type-contents recurring-counts recurring-data-counts)}\\

\textbf{Goals:} \\
\texttt{(and} \\
\texttt{ (done-query query1)}\\
\texttt{ (done-query query2)}\\
\texttt{ (done-query query3)}\\
\texttt{ (in hour-data-counts chat-response)}\\
\texttt{ (in recurring-data-counts chat-response)}\\
\texttt{ (sent chat-response))}\\
\end{tabular}
\smallskip
}

\begin{table}[hbt]
    \centering
    \rowcolors{2}{gray!10}{white}
    \begin{tabular}{>{\bfseries}l l p{8cm}} 
        \toprule
        \rowcolor{gray!20}
        \textbf{Variable} & \textbf{Type} & \textbf{Initial Value/Description} \\
        \midrule
        data-file1              & Data file      & `./genplanx/apps.csv'; file path \\
        dataframe1              & Dataframe       & Empty; Initial dataframe \\
        filtered-2024-dataframe & Dataframe       & Empty; Dataframe for 2024 appointments \\
        filtered-time-dataframe  & Dataframe       & Empty; Dataframe for filter by hour  \\
        filtered-recurring-dataframe & Dataframe   & Empty; Dataframe for filter by recurrence \\
        hour-counts             & Count           & 0 \\
        recurring-counts        & Count           & 0 \\
        hour-data-counts        & Data Contents    & 0 \\
        recurring-data-counts    & Data Contents   & 0 \\
        query1                  & Query           & df[(df[`year'] $== 2024$)] \\
        query2                  & Query           & df[(df[`hour'] $>= 8$) \& (df['hour'] $<= 11$)] \\
        query3                  & Query           & df[df[`isrecurring'] == True] \\
        chat-response            & Response        & Empty \\
        \bottomrule
    \end{tabular}
    \caption{Variables in the appointments filtering configuration with their initial values.}
    \label{table:appointments_filtering_variables}
\end{table}
\begin{figure}[hbt]
    \centering
    \fbox{\includegraphics[width=\linewidth]{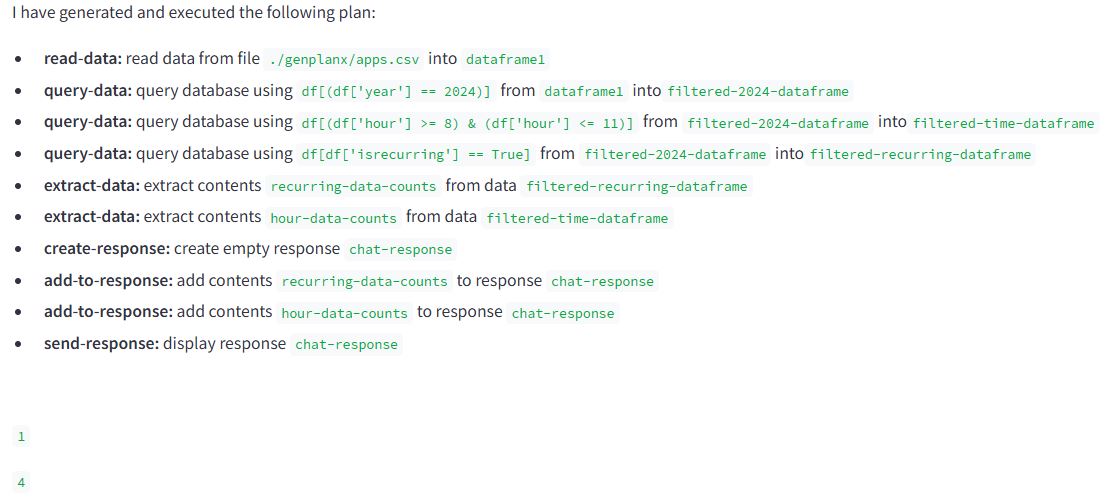}}
    \caption{Appoinments counts created by \genplanx}
    \label{fig:ppt_example2}
\end{figure}

\begin{figure}[hbt]
    \centering
    \begin{tikzpicture}[node distance=1.6cm, auto, thick, 
        state/.style={rectangle, draw, text width=16cm, text centered, minimum height=1.5cm, font=\ttfamily},
        plan/.style={rectangle, draw, text width=10cm, text centered, minimum height=1cm, fill=blue!10, font=\ttfamily},
        goal/.style={rectangle, draw, text width=16cm, text centered, minimum height=1.5cm, fill=green!20, font=\ttfamily},
        every node/.style={font=\ttfamily}]

        \node[state] (init) {\small Initial State: (in dataframe1 data-file1) (available query1) (query-result dataframe1 query1 filtered-2024-dataframe) (available query2) (query-result filtered-2024-dataframe query2 filtered-time-dataframe) (available query3) (query-result filtered-2024-dataframe query3 filtered-recurring-dataframe) (in-data hour-counts filtered-time-dataframe) (data-type-contents hour-counts hour-data-counts) (in-data recurring-counts filtered-recurring-dataframe) (data-type-contents recurring-counts recurring-data-counts)};

        \node[plan, below of=init, yshift=-0.5cm] (step1) {read-data(ai, dataframe1, data-file1)};
        \node[plan, below of=step1] (step2) {query-data(ai, query1, dataframe1, filtered-2024-dataframe)};
        \node[plan, below of=step2] (step3) {query-data(ai, query2, filtered-2024-dataframe, filtered-time-dataframe)};
        \node[plan, below of=step3] (step4) {query-data(ai, query3, filtered-2024-dataframe, filtered-recurring-dataframe)};
        \node[plan, below of=step4] (step5) {extract-data(ai, recurring-counts, filtered-recurring-dataframe, recurring-data-counts)};
        \node[plan, below of=step5] (step6) {extract-data(ai, hour-counts, filtered-time-dataframe, hour-data-counts)};
        \node[plan, below of=step6] (step7) {create-response(ai, chat-response)};
        \node[plan, below of=step7] (step8) {add-to-response(ai, recurring-data-counts, chat-response)};
        \node[plan, below of=step8] (step9) {add-to-response(ai, hour-data-counts, chat-response)};
        \node[plan, below of=step9] (step10) {send-response(ai, chat-response)};

        \node[goal, below of=step10] (goal) {\small Goals: (and (done-query query1) (done-query query2) (done-query query3) (in hour-data-counts chat-response) (in recurring-data-counts chat-response) (sent chat-response))};

        \draw[->] (init) -- (step1);
        \draw[->] (step1) -- (step2);
        \draw[->] (step2) -- (step3);
        \draw[->] (step3) -- (step4);
        \draw[->] (step4) -- (step5);
        \draw[->] (step5) -- (step6);
        \draw[->] (step6) -- (step7);
        \draw[->] (step7) -- (step8);
        \draw[->] (step8) -- (step9);
        \draw[->] (step9) -- (step10);
        \draw[->] (step10) -- (goal);

    \end{tikzpicture}
    \caption{The initial state, goals, and plan for filtering appointments and counting specific types. The initial state and the goals are generated by the LLM. The plan is generated using a classical planner.}
    \label{fig:appointments_filtering_example_plan}
\end{figure}
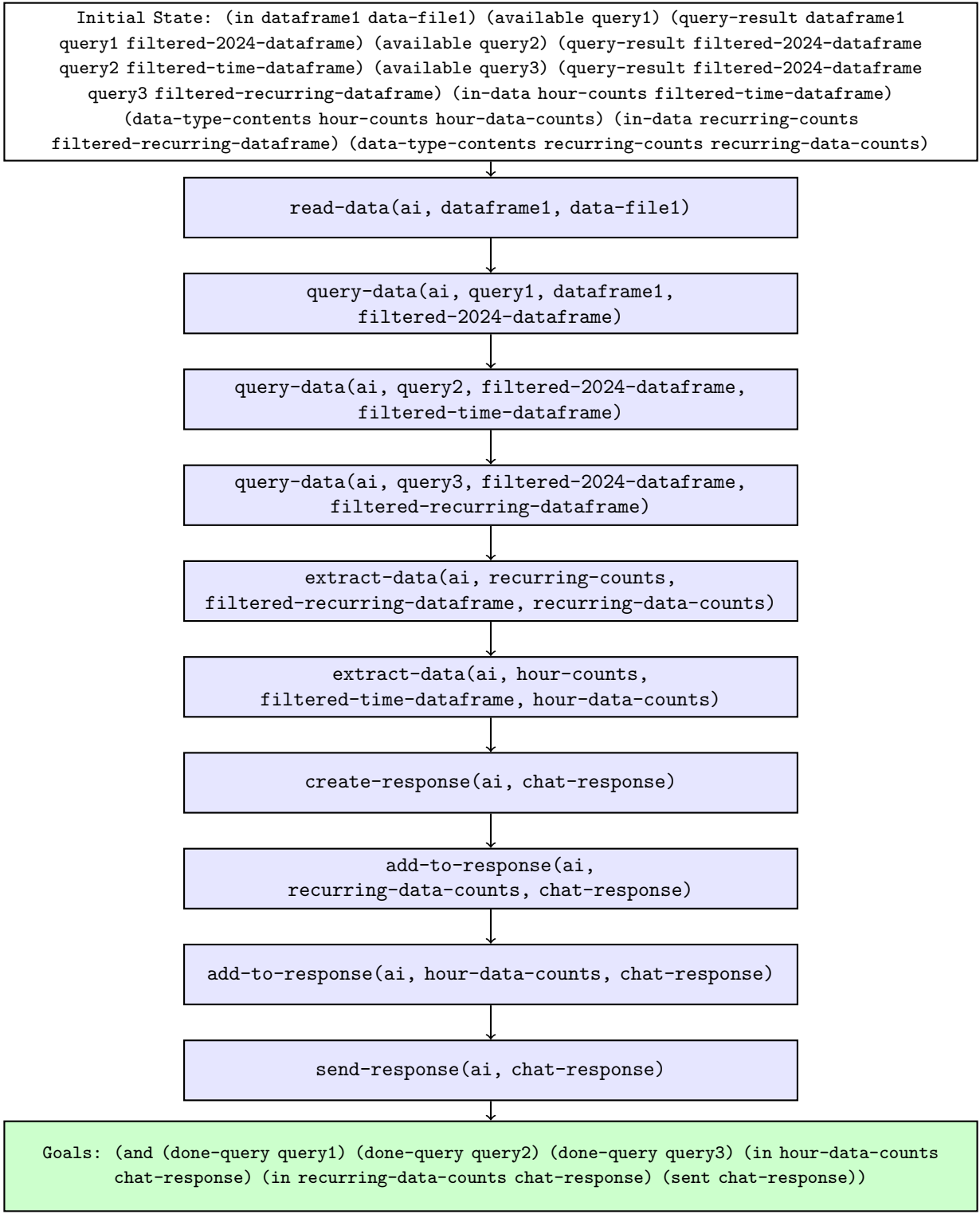

\section{Plan Generation Using LLMs}
\label{sec:llm_for_planning}

In this section, we provide more information on how we use LLM directly to obtain plans in the office domain.

\subsection{Prompting}

To generate a plan using LLM, we provide a prompt describing the actions of the domain and the planning problem.
We provide the description of the actions in natural language. In Table~\ref{table:prompt_slide_example}, we show a prompt with some examples of action descriptions along with the planning problem in the end. We also tested with providing the action descriptions in PDDL and got simular results.

\smallskip
\begin{table}
\begin{tabular}{p{15cm}}

\texttt{
Here are the description of the actions of a planning domain:}

\texttt{The **read-data** action allows an AI agent to access and read data into a dataframe from a database. Once the data is successfully read, it becomes available for further processing or analysis. This action has a parameter called database, indicating that the data is read from this particular database. As a result of executing this action, the total operational cost is increased by the amount database-cost(database).
}

\texttt{
Syntax: read-data(ai-agent, dataframe, database)}
\end{tabular}

\begin{tabular}{p{15cm}}
\texttt{
In the **query-data-optimized** action, an AI agent performs an optimized query on a dataframe, resulting in the availability of the queried data. This action is associated with database-db2, indicating that it leverages optimized querying techniques for efficient data retrieval. The execution of this action increases the total cost by 2 units.}

\texttt{Syntax: query-data-optimized(ai-agent, query, dataframe, data)}

\end{tabular}

\begin{tabular}{p{15cm}}
\texttt{
The **query-data-basic** action allows an AI agent to execute a basic query on a dataframe, making the queried data available. This action is linked to database-db1, suggesting that it uses standard querying methods. As a result, the total cost is increased by 5 units, reflecting the resources consumed during the process.}

\texttt{Syntax: query-data-basic(ai-agent, query, dataframe, data)}
\end{tabular}

\begin{tabular}{p{15cm}}
\texttt{
Through the **create-graph** action, an AI agent can generate a new graph, making it available for visualization or analysis. This action is essential for data representation and incurs a total cost increase of 1 unit, reflecting the resources used in creating the graph.}

\texttt{Syntax: create-graph(ai-agent, graph)}

\end{tabular}

\begin{tabular}{p{15cm}}
\texttt{
The **add-to-graph** action enables an AI agent to incorporate data into a graph, using a reference column from a dataframe. This action is designed to enhance graph content and incurs a total cost increase of 1 unit, reflecting the effort involved in updating the graph.}

\texttt{Syntax: add-to-graph(ai-agent, dataframe, graph, column, referenced-dataframe)}

\end{tabular}

\begin{tabular}{p{15cm}}
\texttt{
In the **create-slide** action, an AI agent creates a new slide with a title in a presentation. This action is crucial for building presentations and incurs a total cost increase of 1 unit, reflecting the resources used in creating the slide.}

\texttt{Syntax: create-slide(ai-agent, slide, presentation, title)}

\end{tabular}

\begin{tabular}{p{15cm}}
\texttt{
The **create-presentation** action allows an AI agent to generate a new presentation, making it available for use. This action is crucial for organizing and delivering information and incurs a total cost increase of 1 unit, reflecting the effort involved in creating the presentation.}

\texttt{Syntax: create-presentation(ai-agent, presentation)}

\end{tabular}

\begin{tabular}{p{16cm}}
\texttt{
Read data and generate a barchart from `balance' against reference column `year’. 
The available databases to read from are db1 with cost of reading 1 and db2 with cost of reading 3. 
db1 supports basic query. db2 supports optimized query. 
Create a slide with bar chart with title `Balance over years', and add it to a presentation. Save the presentation on file genplanx/graph.pptx.
}

\texttt{
Using the above domain model, generate a plan of lowest total cost. The plan should look like
[action1(param11, param12, ….), action2(param21, param22, …), … ]}

\end{tabular}
\caption{Prompt provided to the LLM to generate a plan.}
\label{table:prompt_slide_example}
\end{table}
\subsection{Results of using LLMs for planning}

A summary of the correctness and optimality of the plans generated by the LLMs is shown in Figure~\ref{tab:performance}.
\begin{table}[h!]
\centering
\begin{tabular}{|p{2cm}|p{5 cm}|c|c|c|c|}
\hline
\textbf{LLM} & \textbf{Prompt for Planning} & \multicolumn{2}{c|}{\textbf{Correct Plan}} & \multicolumn{2}{c|}{\textbf{Optimal Plan}} \\ \cline{3-6}
 &  & No hints & With hints & No hints & With hints \\ \hline
GPT4O & Domain and problem in natural language & 4/5 & 5/5 & 0/5 & 5/5 \\ \cline{2-6}
 \hline
 o3-mini & Domain and problem in natural language & 3/5 & 5/5 & 2/5 & 5/5 \\ \cline{2-6}
 \hline
\end{tabular}
\caption{Results of using LLM for planning.}
\label{tab:performance}
\end{table}

Some plans generated by the LLMs include:

\begin{enumerate}
    \item \textbf{Plan 1 (Incorrect):}
    \begin{itemize}
        \item \texttt {read-data(ai-agent, dataframe1, db1)}
        \item \texttt {create-graph(ai-agent, bar-chart1)}
        \item \texttt {add-to-graph(ai-agent, dataframe1, bar-chart1, balance, dataframe1)}
        \item \texttt {create-presentation(ai-agent, presentation1)}
        \item \texttt {create-slide(ai-agent, slide1, presentation1, title1)}
        \item \texttt {add-to-slide-basic(ai-agent, bar-chart1, slide1, presentation1)}
        \item \texttt {generate-presentation(ai-agent, presentation1, presentation-file1)}
    \end{itemize}
    \item \textbf{Plan 2 (Correct but suboptimal):}
    \begin{itemize}
        \item \texttt {read-data(ai-agent, dataframe1, db1)}
        \item \texttt {query-data-basic(ai-agent, query1, dataframe1, filtered-dataframe, db1)}
        \item \texttt {query-data-basic(ai-agent, query2, dataframe1, reference-dataframe, db1)}
        \item \texttt {create-graph(ai-agent, bar-chart1)}
        \item \texttt {add-to-graph(ai-agent, filtered-dataframe, bar-chart1, \\ reference1, reference-dataframe)}
        \item \texttt {create-presentation(ai-agent, presentation1)}
        \item \texttt {create-slide(ai-agent, slide1, presentation1, title1)}
        \item \texttt {add-to-slide-basic(ai-agent, bar-chart1, slide1, presentation1)}
        \item \texttt {generate-presentation(ai-agent, presentation1, presentation-file1)}
    \end{itemize}
    \item \textbf{Plan 3 (optimal):}
    \begin{itemize}
        \item \texttt {create-presentation(ai, presentation1)}
        \item \texttt {read-data(ai, dataframe1, db2)}
        \item \texttt {create-slide(ai, slide1, presentation1, title1)}
        \item \texttt {create-graph(ai, bar-chart1)}
        \item \texttt {query-data-optimized(ai, query1, dataframe1, filtered-dataframe, db2)}
        \item \texttt {query-data-optimized(ai, query2, dataframe1, reference-dataframe, db2)}
        \item \texttt {add-to-graph(ai, filtered-dataframe, bar-chart1, reference1, \\reference-dataframe)}
        \item \texttt {add-to-slide-basic(ai, bar-chart1, slide1, presentation1)}
        \item \texttt {generate-presentation(ai, presentation1, presentation-file1)}
    \end{itemize}
\end{enumerate}

The hints provided to the LLM to arrive at the optimal plan mainly include:
\begin{itemize}
    \item Are you sure this is the optimal plan?
    \item Try again.
    \item You are missing the query step / Query action is missing.
    \item extract-data cannot be used for querying.
    \item Are you using the right database? / Make sure you are using the right database.
\end{itemize}
\end{document}


\appendix \section*{Appendix A -}

The actions defined in the \texttt{assistant} domain are the following:

\begin{enumerate}
    \item \texttt{read-data}: reads data from a data file.
    
    \item \texttt{connect-api}: connects to an API to extract data based on a column and text.
    
    \item \texttt{read-pdf}: reads data from a PDF file.
    
    \item \texttt{read-word}: reads text from a Word file.
    
    \item \texttt{save-pdf}: saves text into a PDF file.
    
    \item \texttt{query-data}: executes a query on a dataframe to filter data.
    
    \item \texttt{extract-data}: extracts data contents from a dataframe.
    
    \item \texttt{create-data}: creates a new dataframe with a single row using the columns from an existing dataframe.
    
    \item \texttt{delete-data}: deletes rows from a dataframe resulting in a new dataframe.
    
    \item \texttt{modify-data}: modifies a dataframe resulting in a new dataframe.
    
    \item \texttt{add-value}: adds a value to a cell in a dataframe.
    
    \item \texttt{modify-row}: modifies a row in a dataframe.
    
    \item \texttt{merge-data}: merges two dataframes into a third one.
    
    \item \texttt{find-info}: finds information based on a text query, making the result available.
    
    \item \texttt{create-graph}: creates a new graph.
    
    \item \texttt{add-to-graph}: 
    
    \item \texttt{create-slide}: creates a new slide with a title in a existing presentation.
    
    \item \texttt{add-to-slide}: adds a graph to a slide in a presentation.
    
    \item \texttt{add-text-to-slide}: adds text to a slide in a presentation.
    
    \item \texttt{add-table-to-slide}: adds a datatable to a slide in a presentation.
    
    \item \texttt{create-presentation}: creates a new presentation.
    
    \item \texttt{contents-in-presentation}: links data contents with a graph and slide in a presentation.???
    
    \item \texttt{generate-presentation}: generates a presentation file from a presentation.???

    \item \texttt{ask-llm}: asks a language model a question and receives an answer.
    
    \item \texttt{search-web}: performs a web search for a query.
    
    \item \texttt{deep-research}: conducts deep research for a query.
    
    \item \texttt{merge-answer}: merges answers from deep research and previous results.
    
    \item \texttt{match-items}: matches items across three files.
\end{enumerate}


%% file: PDDL.tex
\lstdefinelanguage{PDDL}
{
  sensitive=false,    
  morecomment=[l]{;}, 
  alsoletter={:,-},   
  morekeywords={
    define,domain,problem,not,and,or,when,forall,exists,either,
    :domain,:requirements,:types,:objects,:constants,
    :predicates,:action,:parameters,:precondition,:effect,
    :fluents,:primary-effect,:side-effect,:init,:goal,
    :strips,:adl,:equality,:typing,:conditional-effects,
    :negative-preconditions,:disjunctive-preconditions,
    :existential-preconditions,:universal-preconditions,:quantified-preconditions,
    :functions,assign,increase,decrease,scale-up,scale-down,
    :metric,minimize,maximize,
    :durative-actions,:duration-inequalities,:continuous-effects,
    :durative-action,:duration,:condition
  }
}